%% file: main2.tex
\documentclass[runningheads]{llncs}

\usepackage{eccv}

\usepackage{eccvabbrv}

\usepackage{graphicx}
\usepackage{booktabs}
\usepackage[accsupp]{axessibility}  
\usepackage{multirow}
\usepackage{subcaption}  
\usepackage{booktabs}
\usepackage{wrapfig}  
\usepackage[utf8]{inputenc}  
\usepackage[table]{xcolor}
\usepackage[linesnumbered,ruled,vlined]{algorithm2e}  
\usepackage{amsmath,amssymb} 

\usepackage{hyperref}

\begin{document}

\title{StableWorld: Towards Stable and Consistent Long Interactive Video Generation} 

\titlerunning{StableWorld: Stable and Consistent Long Interactive Video Generation}

\author{
Ying Yang\inst{1} \and
Zhengyao Lv\inst{1,2} \and
Yujia Zeng\inst{3} \and
Tianlin Pan\inst{1,4} \and
Haofan Wang\inst{5} \and
Yueming Lyu\inst{1} \and
Binxin Yang\inst{6} \and
Hubery Yin\inst{6} \and
Chen Li\inst{6} \and
Jing LYU\inst{6} \and
Ziwei Liu\inst{7} \and
Chenyang Si\inst{1}\thanks{Corresponding author.}
}

\authorrunning{Y. Yang et al.}

\institute{
PRLab, Nanjing University, China
\and
The University of Hong Kong, Hong Kong SAR, China
\and
University of California, Berkeley, USA
\and
University of Chinese Academy of Sciences, China
\and
LibLib.ai, China
\and
WeChat, Tencent Inc., China
\and
Nanyang Technological University, Singapore
}

\maketitle

\vspace{-1em}
\begin{figure}[h]
    \centering  
    \includegraphics[width=0.97\textwidth]{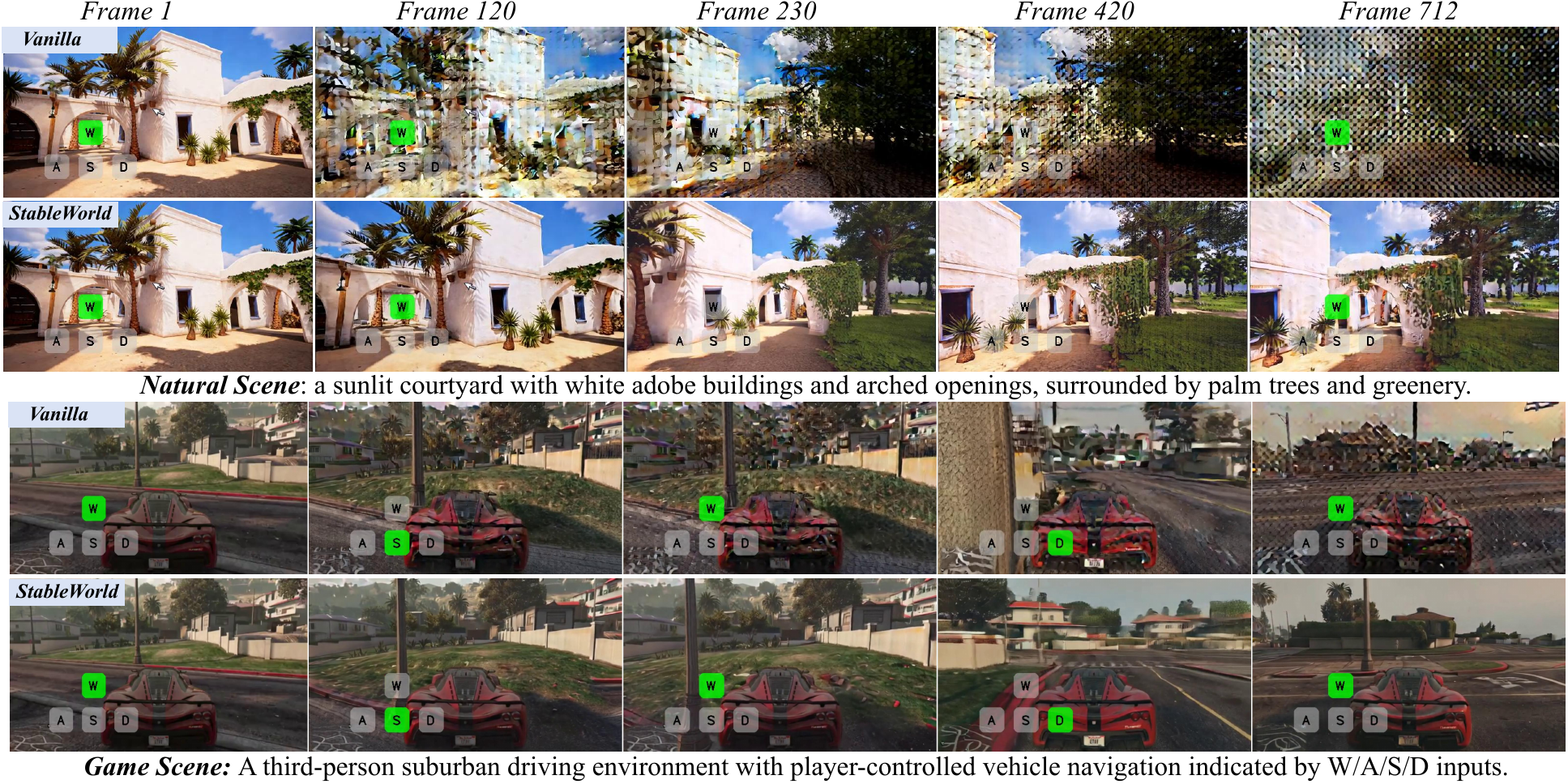}
    \caption{\textit{\textbf{StableWorld}}: producing stable and visually consistent interactive videos across diverse environments such as natural landscapes and game worlds, while preserving continuous motion control and preventing long-term scene drift.}
    \label{fig:teaser}
\end{figure}
\vspace{-3em}

\input{sec/0_abstract}  
\vspace{-0.2em}
\input{sec/1_intro}
\input{sec/2_related}

\input{sec/3_Method}

\input{sec/4_exp}

\section{Conclusion and Future Work}
In this paper, we identify a common problem faced by current interactive video generation models: \emph{scene collapse}. 
Through further analysis, we find that this collapse originates from the inter-frame drift that occurs between adjacent frames within the same scene, which gradually accumulates over time and eventually leads to a large deviation from the original scene. Motivated by this observation, we propose a simple yet effective method — \textbf{StableWorld}, a dynamic frame eviction mechanism that significantly reduces error accumulation while preserving motion consistency.  We evaluate our approach on multiple interactive video generation models, including \textit{Matrix-Game 2.0}, \textit{Open-Oasis}, and \textit{Hunyuan-GameCraft 1.0}. Extensive qualitative and quantitative experiments demonstrate that our method substantially improves visual quality over long horizons and shows strong potential for integration with future world models.

\noindent\textbf{Future Work.} 
In future work, we plan to explore integrating StableWorld into the training process to further extend the length and stability of interactive video generation. 
Moreover, the ability to identify and discard a large number of drifted frames during generation has the potential to reduce training cost while maintaining generation quality and aligns naturally with future extensions toward memory-augmented world models.

\label{sec:con}

\par\vfill\par


%
%
\bibliographystyle{splncs04}
\bibliography{main}
\input{sec/6_appendix}

\end{document}

%% file: sec/0_abstract.tex
\begin{abstract}

In this paper, we explore the overlooked challenge of stability and temporal consistency in interactive video generation, which synthesizes dynamic and controllable video worlds through interactive behaviors such as camera movements and text prompts. Despite remarkable progress in world modeling, current methods still suffer from severe instability and temporal degradation, often leading to spatial drift and scene collapse during long-horizon interactions. To better understand this issue, we initially investigate the underlying causes of instability and identify that the major source of error accumulation originates from the same scene, where generated frames gradually deviate from the initial clean state and propagate errors to subsequent frames. Building upon this observation, we propose a simple yet effective method, \textbf{StableWorld}, a Dynamic Frame Eviction Mechanism. By continuously filtering out degraded frames while retaining geometrically consistent ones, StableWorld effectively prevents cumulative drift at its source, leading to more stable and temporal consistency of interactive generation. Promising results on multiple interactive video models, \eg, Matrix-Game, Open-Oasis, and Hunyuan-GameCraft, demonstrate that StableWorld is model-agnostic and can be applied to different interactive video generation frameworks to substantially improve stability, temporal consistency, and generalization across diverse interactive scenarios.

  \keywords{World Model \and Interactive Video Generation}
\end{abstract}

%% file: sec/1_intro.tex
\section{Introduction}
\label{sec:intro}

\begin{figure}[t]
    \centering
    \begin{minipage}{0.48\linewidth}  
        \centering
        \includegraphics[width=\linewidth,]{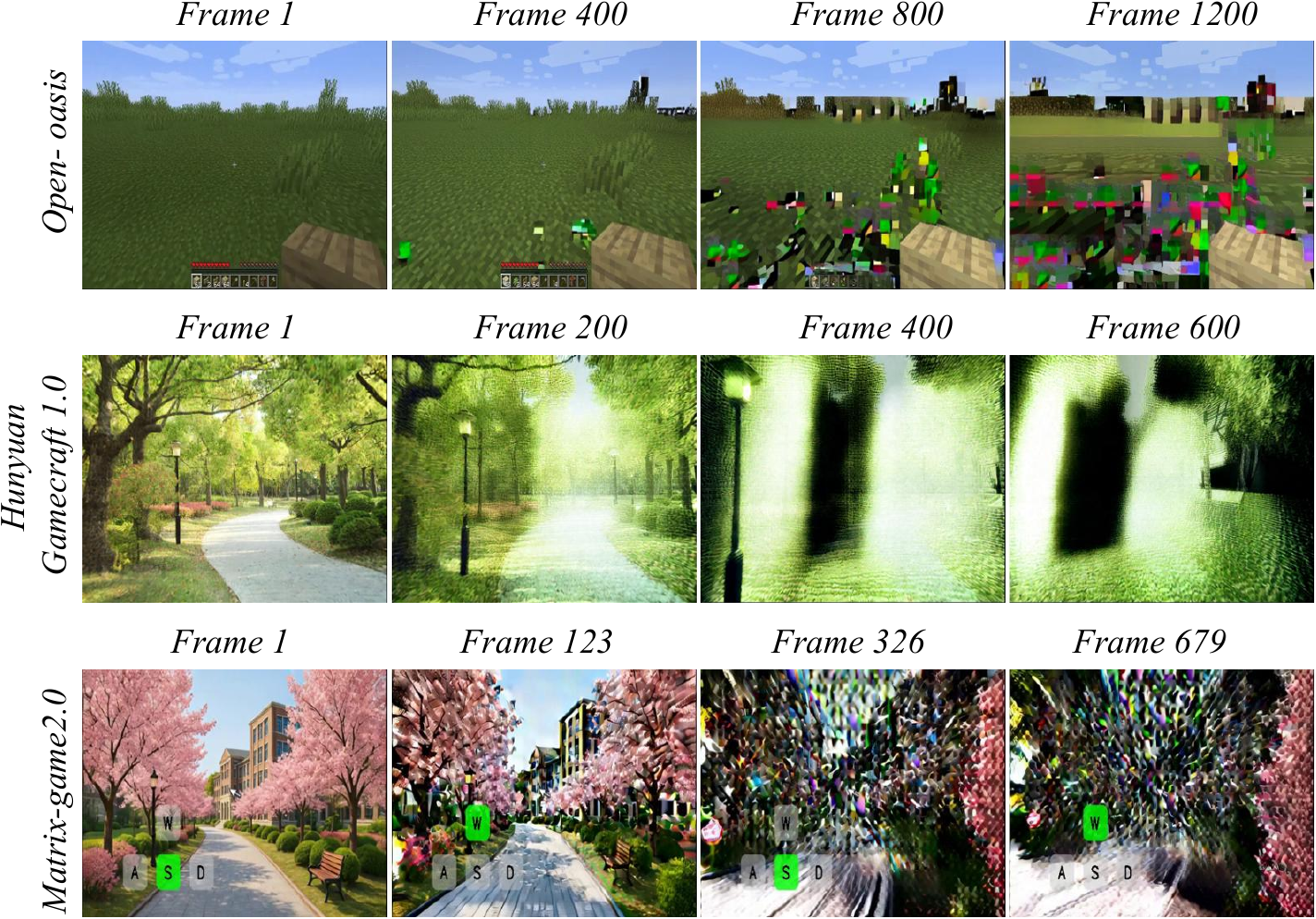}
        \vspace{0.3em}
        \caption{Visualization of progressive scene collapse over time across different world-simulation models.}
        \label{fig:bad_result}
    \end{minipage}
    \hfill  
    \begin{minipage}{0.48\linewidth}
        \centering
        \includegraphics[width=\linewidth,height=4.5cm,]{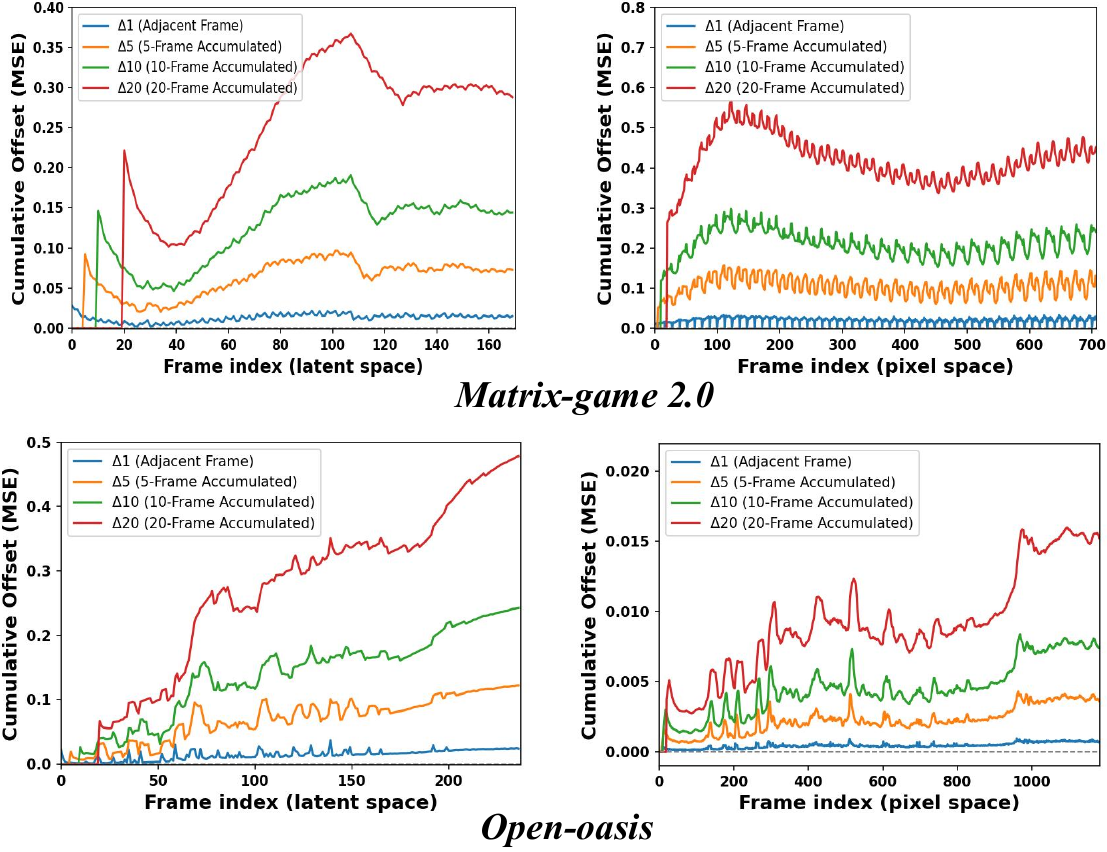}
        \caption{Accumulated frame-to-frame drift in \textbf{Matrix-Game 2.0} and \textbf{Open-Oasis}.}
        \label{fig:trend}
    \end{minipage}
\end{figure}

Recent advances in video generation~\cite{kong2024hunyuanvideo,wan2025wan,wang2025lavie,teng2025magi} have achieved remarkable success in learning rich spatio-temporal knowledge from large-scale videos. Building upon these capabilities, recent studies~\cite{ha2018recurrent,oh2015action,bruce2024genie,decart2024oasis,zhang2025matrix,he2025matrix,li2025hunyuan,guo2025mineworld,yu2025context,xiao2025worldmem} have leveraged video generation models to understand real-world interactions and predict environment dynamics conditioned on actions, extending them toward the paradigm of world modeling. However, as shown in Fig.~\ref{fig:bad_result}, current world models commonly suffer from a phenomenon of progressive scene collapse as time elapses, particularly in static or slowly changing environments. Hence, ensuring long-term stability and temporal consistency across diverse actions and scenes, especially in generating long interactive videos without scene collapse, remains a fundamental yet insufficiently explored challenge in world modeling.

In this paper, we delve into the phenomenon of degradation in world models. For a comprehensive analysis, we begin with the simplest form of interaction, \ie static interaction, by generating long static video sequences to observe how scene stability evolves over time. To this end, we measure inter-frame mean squared error (MSE) distances to quantify how frame discrepancies accumulate as the sequence progresses. As illustrated in Fig.~\ref{fig:trend}, while consecutive frames differ only slightly, these subtle deviations accumulate over time, leading to a noticeable drift from the initial clean state. As the drift accumulates over time, it manifests as visible inconsistencies and eventually leads to progressive scene collapse. These observations suggest that error accumulation within the same scene, even under minimal interaction, is a key factor contributing to instability in current world models. 

\begin{figure*}[t]
    \centering
    \includegraphics[width=\textwidth]{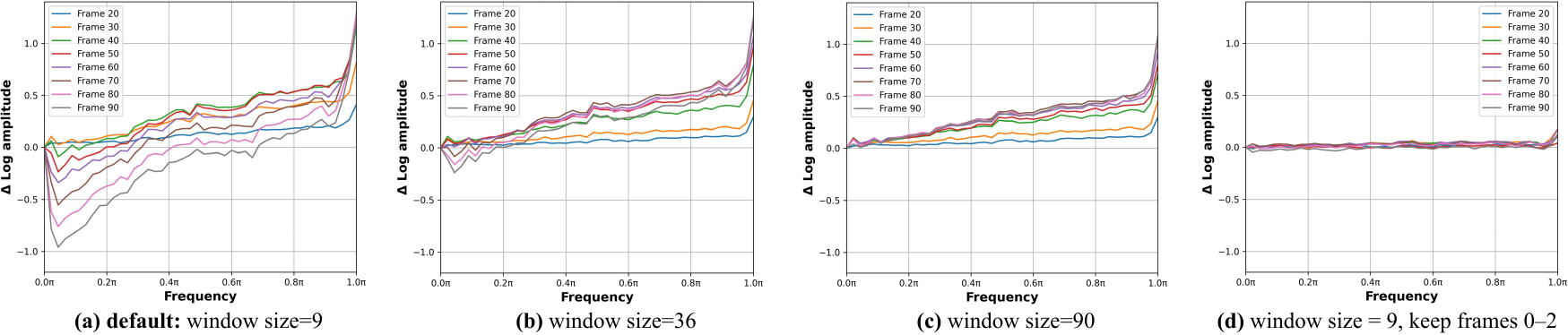}
\caption{
Frequency amplitude difference between the anchor frame and different target frames. 
(a) Default setting: window size = 9. 
(b) window size = 36. 
(c) window size = 90. 
(d) window size = 9 with the first clean 3 frames retained.
}

    \label{fig:fig3}
\end{figure*}

Based on this observation, we hypothesize that using frames with smaller accumulated drift as historical references can provide a more stable foundation for subsequent frame generation. To verify this hypothesis, we analyze the impact of enlarging the KV-cache window size, which allows the model to access earlier and potentially cleaner frames during generation. As shown in Fig.~\ref{fig:fig3}, increasing the window size effectively mitigates the degradation phenomenon by reducing frequency amplitude fluctuations between each target frame and the first frame. This finding suggests that incorporating temporally cleaner frames can suppress cumulative errors and stabilize long-horizon generation. Further analysis reveals that the observed stabilization when enlarging the history window primarily arises from retaining several clean early frames within the reference buffer.

\begin{figure*}[t]
    \centering
    \includegraphics[width=\textwidth]{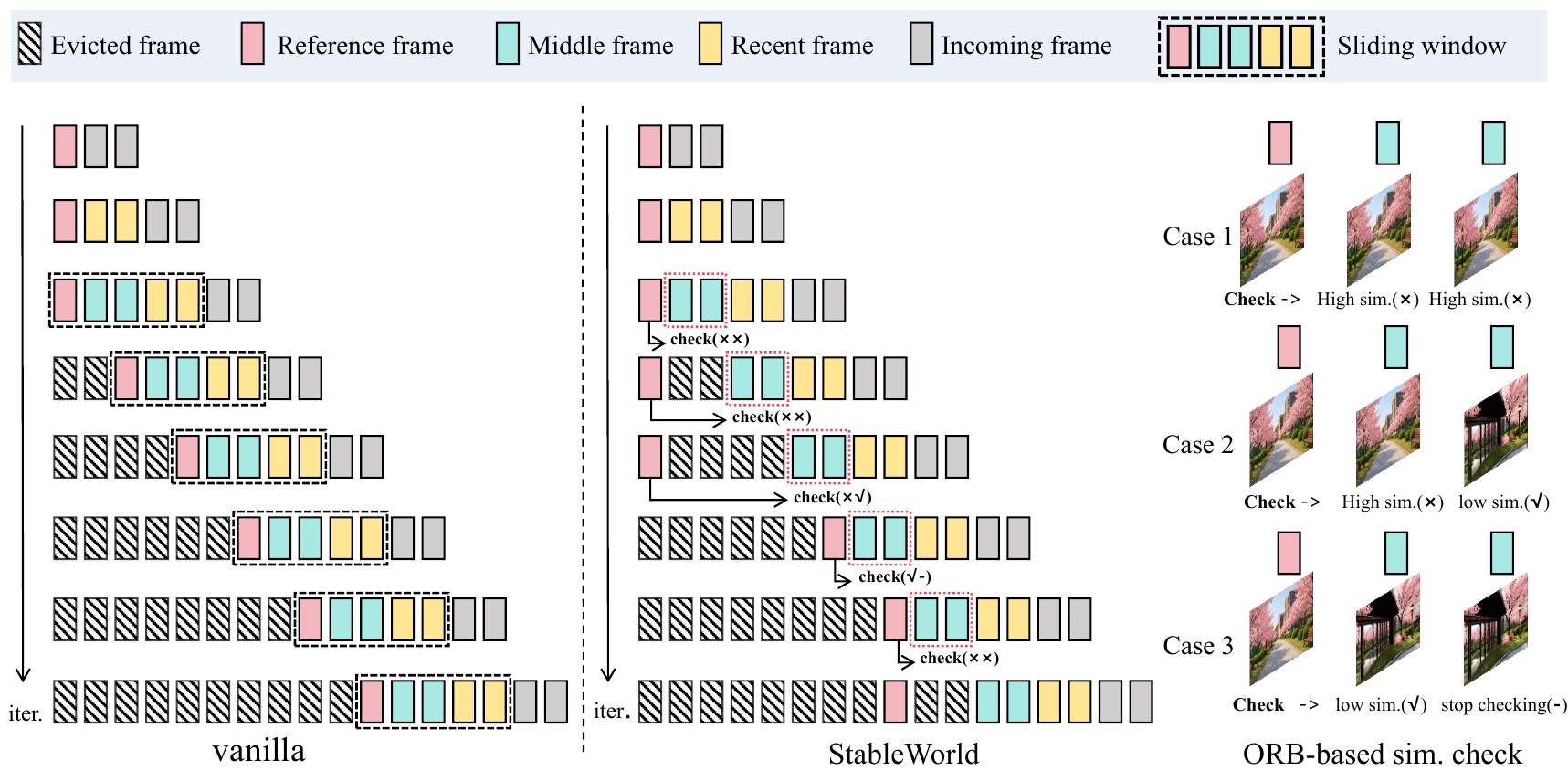}
    \caption{
Comparison of sliding-window updates. 
When the window slides, the vanilla model simply discards the oldest frames, causing errors to accumulate over iterations. 
In contrast, \textbf{StableWorld} continuously filters out degraded frames while maintaining geometrically consistent ones, resulting in more stable and consistent interactive video generation. 
    }
    \label{fig:overall}
\end{figure*}

With these revelations as our backdrop, we introduce \textbf{StableWorld}, a simple yet effective framework for building an interactive world model capable of generating stable and temporally consistent videos, as shown in Fig.~\ref{fig:overall}. The core design of StableWorld is Dynamic Frame Eviction Mechanism to dynamically maintain a clean and reliable historical frame representations in the sliding window for current-frame generation. When the window needs to slide as new frames are appended, some historical frames must be evicted. We always retain the most recent frames, since they are critical for preserving local motion continuity and visual smoothness. For earlier frames, we designate the earliest, least-degraded frames in the window as the reference frames, and evaluate geometric consistency between it and intermediate frames using a viewpoint-overlap score computed via ORB + RANSAC~\cite{rublee2011orb,fischler1981random}. If multiple intermediate frames share a similar viewpoint with the reference, they are regarded as redundant and potentially drifted, and are selectively evicted to prevent further error accumulation. \textbf{By continuously filtering out degraded frames while retaining geometrically consistent ones}, StableWorld effectively suppresses cumulative errors while preserving adaptability to large motions and scene transitions, enabling stable and consistent interactive video generation.

We conduct a comprehensive experimental evaluation of our approach. 
The results demonstrate that \textbf{StableWorld} significantly reduces cumulative drift while maintaining motion continuity across diverse scenarios, as shown in Fig.~\ref{fig:teaser}.
Our contributions are summarized as follows:

\begin{itemize}
  \item We identify the root cause of instability in long interactive world modeling: small drifts accumulate within the same scene, eventually leading to overall scene collapse.

\item We propose a simple yet effective method \textbf{StableWorld}, 
a dynamic frame eviction mechanism that effectively prevents error accumulation at its source while maintaining motion continuity.
  
  \item We validate our method across multiple interactive world models—Matrix-Game 2.0, Open Oasis, and Hunyuan-GameCraft 1.0—under diverse conditions (static scenes, small/large motions, and significant viewpoint changes). Extensive results show consistent improvements in stability, long-term consistency, and generalization across various interactive scenarios.
\end{itemize}

%% file: sec/2_related.tex
\section{Related work} 
\label{sec:formatting}

\textbf{Video Generation Models.} 
With the rapid development of diffusion models~\cite{ho2020denoising,song2020denoising,ho2022classifier,peebles2023scalable,rombach2022high,batifol2025flux}, video generation has also made remarkable progress in producing high-quality and temporally coherent videos~\cite{hong2022cogvideo,yang2024cogvideox,kong2024hunyuanvideo,wan2025wan,bao2024vidu,wang2025lavie,brooks2024video,zhang2025packing,zheng2024open,lin2024open}. 
These methods typically extend image diffusion into the spatiotemporal domain, enabling globally consistent visual appearance across multiple frames. Meanwhile, autoregressive models have achieved significant advances in both image~\cite{tian2024visual,sun2024autoregressive,han2025infinity,zhou2024transfusion,yu2023language} and video generation~\cite{henschel2025streamingt2v,teng2025magi,huang2025self,cui2025self,liu2025rolling,jin2024pyramidal,yin2025slow,wang2024loong,liu2025infinitystar} due to their strong scalability and controllability. 
By generating videos in a step-by-step manner conditioned on previously produced tokens, autoregressive models naturally support long-horizon temporal modeling and interactive generation. 

\noindent\textbf{Interactive World Simulation.} 
Generating interactive environments can be regarded as a form of world simulation, often referred to as \emph{world models}~\cite{ha2018recurrent,oh2015action,bruce2024genie}, which predict the next state in an autoregressive manner conditioned on the current state and action.
Recently, the emergence of high-quality world models built upon powerful video generation techniques has led to rapid progress~\cite{decart2024oasis,zhang2025matrix,he2025matrix,li2025hunyuan,guo2025mineworld,yu2025gamefactory,yang2023learning}. Some studies~\cite{decart2024oasis,li2025hunyuan,guo2025mineworld} focus on improving action following and maintaining temporal continuity, while others~\cite{yu2025context,xiao2025worldmem} aim to enhance memory consistency to ensure that the scene remains coherent under the same viewpoint across subsequent generations. These efforts collectively improve interactive controllability and short-term coherence in dynamic environments.

\noindent\textbf{Long Video Generation.} 
While recent video generation models have achieved impressive visual quality, they still struggle to produce long videos, typically limited to around 5 seconds~\cite{wan2025wan,kong2024hunyuanvideo,yang2024cogvideox,bao2024vidu}, due to prohibitive computational costs and cumulative drift. A variety of approaches have been proposed for extending long-context generation. Methods such as~\cite{li2025hunyuan,team2025longcat} generate videos chunk by chunk, where each new segment is conditioned on the previously generated one. Although this strategy enables longer temporal horizons, it introduces high training cost and often causes motion discontinuities at chunk boundaries. Frame-packing~\cite{zhang2025packing} produces keyframes across different stages and interpolates intermediate frames accordingly. While this constrains error propagation between adjacent keyframes, it inevitably limits motion flexibility and generative diversity. Other approaches, including diffusion forcing~\cite{chen2024diffusion} and self-forcing~\cite{huang2025self}, simulate autoregressive degradation patterns during training to reduce the train–inference distribution gap and improve long-term temporal stability. However, these methods still suffer from error accumulation during inference. Moreover, these approaches remain largely confined to conventional Text-to-Video (T2V) or Image-to-Video (I2V) settings, leaving long interactive video generation insufficiently addressed.

%% file: sec/3_Method.tex
\section{Methodology}
\label{method}
\subsection{Preliminary}



\textbf{Video Generation Models.}  
Video generation models typically adopt a \emph{full-sequence} generation approach, which generates all frames at once from noise under a given condition $\boldsymbol{c}$. 
Formally, the generation process can be defined as:
\begin{equation}
    p_{\theta}(\boldsymbol{x}_{k}^{t} \mid \boldsymbol{x}_{k}^{t+1},  \boldsymbol{c})
    = \mathcal{N}\bigl(\boldsymbol{x}_{k}^{t}; \mu_{\theta}(\boldsymbol{x}_{k}^{t+1}, \boldsymbol{c}, t), \sigma_{t}^{2}\mathbf{I}\bigr),
    \label{eq:full_seq}
\end{equation}
where $\boldsymbol{x}_{k}^{t}$ denotes the $t$-th denoising step of the $k$-th frame, and $0 \leq k \leq K$ with $K$ being the total number of generated frames.  
All frames share the same noise variance $\sigma_t$ at each timestep $t$, following a unified noise schedule.  
Although this approach achieves high-quality results, modeling the entire sequence in a single forward pass incurs a high computational cost and is not suitable for real-time interactive scenarios.

\noindent\textbf{Interactive Video Generation.}
In contrast to full-sequence models, interactive video generation adopts an autoregressive paradigm, where each frame $\boldsymbol{x}_{k}$ is generated conditioned on a subset of historical frames and the current action $a_{k}$. This conditional generation is expressed as $p_{\theta}(\boldsymbol{x}_{k} \mid \boldsymbol{x}_{\{1:k-1\}}^{\text{subset}}, a_{k})$,
where $\boldsymbol{x}_{\{1:k-1\}}^{\text{subset}}$ denotes the selected reference frames maintained in the memory buffer, and $a_{k}$ represents the user-issued or agent-driven action at step $k$.
This paradigm allows the model to generate frames sequentially in response to user actions, enabling real-time interaction and dynamic scene control.

Most recent approaches further combine diffusion and autoregressive paradigms: 
diffusion models are used for \emph{intra-frame} denoising, while autoregression captures \emph{inter-frame} temporal dependencies.  
Formally, the overall generation process can be expressed as:
\vspace{-0.5em}

\begin{equation}
    p_{\theta}\left(
    \boldsymbol{x}_{k}^{t} \mid \boldsymbol{x}_{k}^{t-1}, \boldsymbol{x}_{\{1:k-1\}}^{\text{subset}}, a_{k}
    \right)
    = \mathcal{N}\left(
    \boldsymbol{x}_{k}^{t}; \mu_{\theta}\left(
    \boldsymbol{x}_{k}^{t-1}, \boldsymbol{x}_{\{1:k-1\}}^{\text{subset}}, a_{k}, t
    \right), \sigma_{t}^{2} \mathbf{I}
    \right)
\label{eq:diff_ar}
\end{equation}
\vspace{-0.05em}
where $\boldsymbol{x}_{k}^{t}$ denotes the $k$-th frame at diffusion timestep $t$. At each diffusion step $t$, the model denoises $\boldsymbol{x}_{k}^{t-1}$ into $\boldsymbol{x}_{k}^{t}$ conditioned on previously generated frames $\boldsymbol{x}_{\{1:k-1\}}^{\text{subset}}$ and the current action $a_{k}$.  
This formulation integrates spatial denoising within each frame and temporal dependencies across frames, enabling high-quality and real-time interactive video generation.

\subsection{The Reason for Scene Collapse}

\begin{figure}[t]
    \centering
    \begin{minipage}{0.47\linewidth}  
        \centering
        \includegraphics[width=\linewidth]{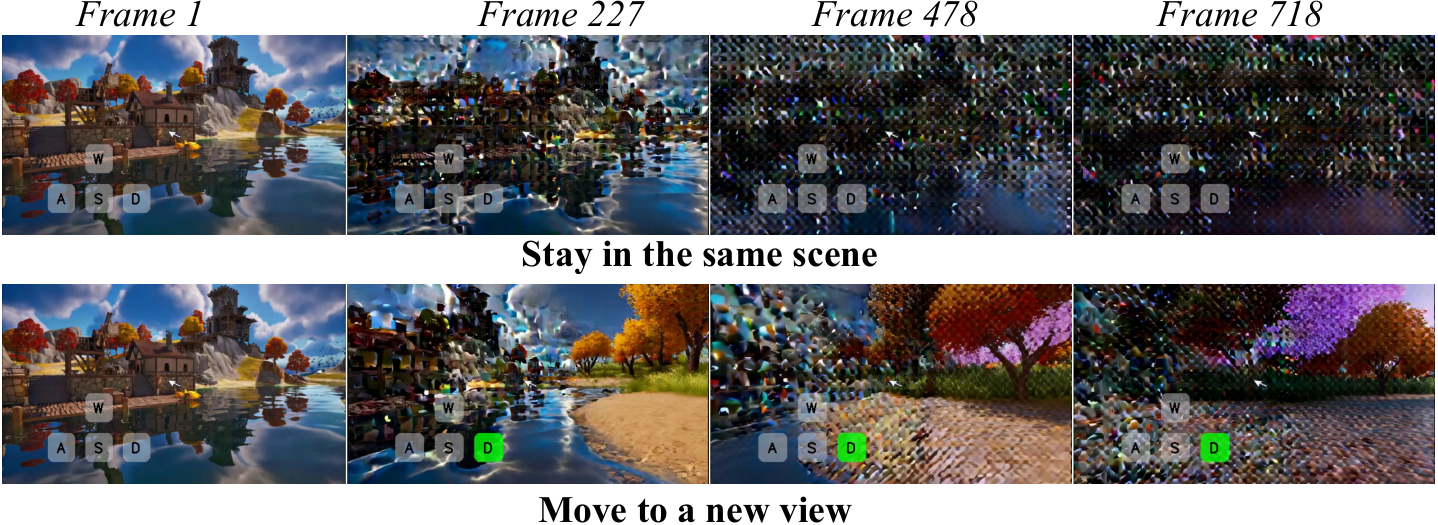}
        \caption{Drift Accumulation and Propagation Across Views.The first row remains in the same scene, while the second row moves to a new viewpoint.}
        \label{fig:fre}
    \end{minipage}
    \hfill  
    \begin{minipage}{0.49\linewidth}  
        \centering
        \includegraphics[width=\linewidth]{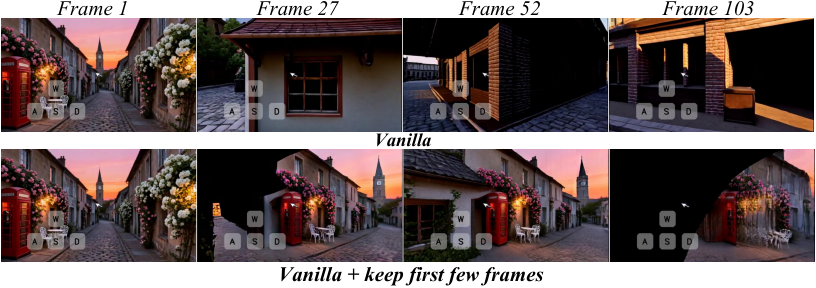}
        \caption{Comparison under scene switching settings. The first row shows the baseline vanilla model, while the second row keeps the first three frames unchanged.}
        \label{fig:compare1}
    \end{minipage}
\end{figure}

Although interactive video generation models can produce coherent short-term sequences, they still tend to exhibit progressive \emph{scene collapse} during long-duration generation, particularly when the scene remains highly similar for an extended period (Fig.~\ref{fig:bad_result}). Furthermore, the drift accumulated within the same scene can propagate to newly generated views. As shown in Fig.~\ref{fig:fre}, the first row remains in the same scene, while the second row moves to a new viewpoint. The drift accumulated in previous views affects the generation of subsequent views, eventually leading to complete scene collapse. 
These observations suggest that the collapse is largely caused by how visual information is preserved and propagated within the same scene over time.




To understand this behavior, we measure the inter-frame mean squared error (MSE) distances to quantify how frame discrepancies change as the sequence progresses within a single static scene, as shown in Fig.~\ref{fig:trend}. 
The left two plots visualize inter-frame drift at different intervals (1, 5, 10, 20) in the latent space. 
We observe that although adjacent frames exhibit only minor discrepancies, these small drifts gradually accumulate as the sequence extends. 
Frames compared over larger intervals (e.g., 10 or 20) show significantly greater drift.
Since such deviations are already present in the latent space, the pixel space also exhibits a similar drift pattern, as shown in the right two plots, which ultimately manifests as visual inconsistency and scene collapse(see Fig.~\ref{fig:bad_result}). 
These observations demonstrate that \textbf{drift within a single scene accumulates and propagates over time, ultimately leading to global scene collapse}.

Based on this observation, we hypothesize that using frames with smaller accumulated drift can provide more reliable references for subsequent frame generation. 
To verify this hypothesis, we enlarge the KV-cache window size, allowing the model to access cleaner frames, as shown in Fig.~\ref{fig:fig3}. 
We examine how the frequency amplitude difference between each target frame and the first frame varies under different history window sizes. 
With the default setting (window size = 9, panel (a)), large variations appear across all frequency bands. 
As the window size increases to 36 (panel (b)) and 90 (panel (c)), the overall fluctuation slows down, indicating a partial reduction of error accumulation. 
However, this improvement comes at the cost of higher computational overhead and slower generation, which limits its practicality.


Further analysis reveals that the observed stabilization from a larger window primarily arises from retaining several clean early frames in the reference buffer. 
For example, in panel (d), preserving a few reliable early frames within a fixed-size window leads to notably more stable generation, with later frames showing minimal drift relative to the first frame. 
This finding highlights that \textbf{the quality and preservation of early clean frames play a crucial role in mitigating cumulative errors}. 
Nevertheless, when large motions or drastic scene transitions occur, always retaining the initial frames becomes restrictive.
As shown in Fig.~\ref{fig:compare1}, although both settings follow the same action instructions, the one that rigidly preserves early frames fails to switch to a new scene, suggesting that excessive retention hinders scene transitions. 
To simultaneously mitigate cumulative drift and preserve the flexibility to generate new scenes, we introduce \textbf{StableWorld}, a simple yet effective framework built upon a \textit{Dynamic Frame Eviction Mechanism} (see Fig.~\ref{fig:overall}), which will be described in the following section.

\subsection{Dynamic Frame Eviction via ORB-based Geometric Similarity}
To determine whether a scene transition occurs, we employ ORB~\cite{rublee2011orb} feature matching combined with RANSAC-based geometric verification to measure inter-frame similarity.
When there aren't explicit camera extrinsic parameters in the inference stage, ORB provides an alternative that can produce fast and rotation-invariant local features, making it well-suited for detecting geometric consistency under small camera motions.
By integrating this similarity estimation with a dynamic frame eviction strategy, we \textbf{continuously filter out degraded frames while retaining geometrically consistent ones}, thereby effectively preventing error accumulation across dynamic scenes.

Specifically, when the window needs to slide, 
some frames must be evicted. For simplicity, we assume that each frame corresponds to one token, and one token is generated at each iteration.
Let $\{\boldsymbol{L}_0, \boldsymbol{L}_1, \ldots, \boldsymbol{L}_{N-1}\}$ denote the latent-space tokens within the window, and 
$\{P_0, P_1, \ldots, P_{N-1}\}$ denote their corresponding pixel-space frames, 
where $N$ is the window size. 
The earlier frames in the window are defined as $\{P_0, P_1, \ldots, P_K\}$ with $K < N-1$. 
Here, $P_0$ is treated as the reference frame, while $\{P_1, \ldots, P_K\}$ are referred to as the middle frames.
At each update step, one new frame is generated, and one old frame is evicted accordingly.

We determine which frame should be evicted using the following strategy. 
First, we measure geometric similarity by extracting ORB features from the reference frame $P_0$ and a middle frame $P_k$ ($k \ge 1$). 
Let $\{\boldsymbol{d}_i^{(0)}\}_{i=1}^{M_0}$ and $\{\boldsymbol{d}_j^{(k)}\}_{j=1}^{M_k}$ denote the sets of ORB descriptors
extracted from $P_0$ and $P_k$, respectively,
where $M_0$ and $M_k$ are the numbers of detected features in each frame.
Candidate correspondences $G$ are obtained by nearest-neighbor matching in descriptor space, 
followed by Lowe’s ratio test:
\begin{equation}
    G = \{(i,j) \mid \| \boldsymbol{d}_i^{(0)} - \boldsymbol{d}_j^{(k)} \| < \tau \}, \quad g = |G|,
\label{G}
\end{equation}
where $\tau$ is the ratio-test threshold used to filter ambiguous matches, and $g = |G|$ 
denotes the number of surviving correspondences.

The matches in $G$ are then verified using RANSAC with both Homography ($\mathbf{H}$) and Fundamental matrix ($\mathbf{F}$) models
to enforce geometric consistency:
\begin{align}
    \mathcal{I}_{H} &= \{ (i,j) \in G \mid \text{SampsonErr}_{\mathbf{H}}(i,j) \leq \epsilon \}, \\
    \mathcal{I}_{F} &= \{ (i,j) \in G \mid \text{SampsonErr}_{\mathbf{F}}(i,j) \leq \epsilon \},
\label{Sam}
\end{align}
where $\text{SampsonErr}_{\mathbf{H}}(i,j)$ and $\text{SampsonErr}_{\mathbf{F}}(i,j)$ denote the Sampson geometric errors 
evaluated under the estimated Homography $\mathbf{H}$ and Fundamental matrix $\mathbf{F}$, respectively, and 
$\mathcal{I}_{H}$ and $\mathcal{I}_{F}$ denote the corresponding inlier correspondence sets. 
$\epsilon$ is a predefined tolerance for inlier determination, with smaller errors indicating better geometric alignment .
We compute the inlier ratios:
\begin{equation}
    r_H = \frac{|\mathcal{I}_{H}|}{g}, \quad r_F = \frac{|\mathcal{I}_{F}|}{g},
\end{equation}
where $|\mathcal{I}_{H}|$ and $|\mathcal{I}_{F}|$ denote the numbers of inlier correspondences under the two models.
The final similarity score is defined as:
\begin{equation}
    s(P_0, P_k) = \max(r_H, r_F).
\end{equation}
If the similarity score $s(P_0, P_k)$ exceeds a predefined threshold $\theta$, 
the check continues iteratively for farther frames ($P_{k+1}, P_{k+2}, \ldots$).
The process stops once geometric similarity drops below $\theta$.  
Finally, if all middle frames satisfy the threshold, the farthest frame $P_K$ is evicted.
Otherwise, the frame directly preceding the first failure (e.g., \(P_{k-1}\)) is evicted.
The detailed algorithmic procedure and implementation settings are provided in Appendix~\ref{Alg}.






%% file: sec/4_exp.tex
\section{Experiments}
\label{exp}

\subsection{Evaluation Metrics.}

To evaluate the effectiveness of our proposed method compared to the baseline, we conduct both qualitative and quantitative experiments across multiple models and scenarios. Given that the three world models were trained on distinct datasets, \textbf{we utilize the official datasets from each model's repository to ensure fairness in evaluation.}
For \textit{Matrix-Game 2.0}~\cite{he2025matrix}, we evaluate 16 different scenes, covering various environments as well as small and large motion actions, resulting in a total of 96 one-minute videos. 
For \textit{Open-Oasis}~\cite{decart2024oasis}, we evaluate 10 different scenes and generate 80 one-minute videos involving diverse motions. 
For \textit{Hunyuan-GameCraft 1.0}~\cite{li2025hunyuan}, we evaluate 16 scenes and generate 64 videos, each lasting 45 seconds.
We use the VBench-Long~\cite{huang2024vbench} benchmark to assess multiple dimensions of video quality and consistency. Further, we also conduct a user study with 20 participants to assess the video quality, temporal consistency, and the motion smoothness in 30 videos of the three models. 
In addition, we also verify the effectiveness of \textit{StableWorld} as an error mitigation strategy in autoregressive video generation, evaluated under the \textbf{Self-Forcing}~\cite{huang2025self}(see Appendix~\ref{sec:more}).

\subsection{Implementation Details}
For \textit{Matrix-Game 2.0}, we set the key-value cache length to $9$ and treat the earliest $6$ frames as earlier frames. 
Since the model generates $3$ frames at each step, we compute the ORB-based similarity for the 3rd and 6th frames to determine which frames to evict.
For \textit{Open-Oasis}, the model generates one frame per step with a default window size of $16$. 
We designate the earliest $12$ frames in the window as earlier frames and evaluate ORB-based similarity for the 1st, 6th, and 12th frames.
For \textit{Hunyuan-GameCraft 1.0}, the model generates $33$ frames at each step, where the newly generated $33$ frames form the new window. 
We compare the 1st, 6th, and 12th frames of the previous window with their corresponding frames in the new window. 
If the similarity is above the threshold, the earlier frames are merged into the new window.
For all models, we set the ORB-based similarity threshold to $0.75$.

\begin{table}[t]
  \centering
  \caption{Quantitative comparison on VBench-Long~\cite{huang2024vbench} across three vanilla models and their StableWorld-enhanced counterparts. 
  Our StableWorld method improves both visual and temporal quality with minimal additional latency.}
  \resizebox{\textwidth}{!}{  
    \begin{tabular}{c|c|ccc|cc|cc}
    \toprule
    \multirow{2}[3]{*}{Method} & \multirow{2}[3]{*}{Latency} & \multicolumn{3}{c|}{Visual Quality} & \multicolumn{2}{c|}{Temporal Quality} & \multicolumn{2}{c}{Physical Understanding} \\
    \cmidrule{3-9}
          &       & Image Quality $\uparrow$ & Aesthetic $\uparrow$ & Dynamic Degree $\uparrow$ & Temporal Flickering $\uparrow$ & Motion Smooth $\uparrow$ & Subject Cons. $\uparrow$ & Background Cons. $\uparrow$ \\ 
    \midrule
    Matrix-Game 2.0 & $\times$1.00 & 63.39 & 38.83 & 56.19 & 95.19 & 97.41 & 96.49 & 97.52 \\ 
    \rowcolor{gray!10} StableWorld + Matrix-Game 2.0 & $\times$1.01 & \textbf{73.52} & \textbf{53.44} & \textbf{56.66} & \textbf{95.96} & \textbf{98.13} & \textbf{97.02} & \textbf{97.74} \\
    \midrule
    Open-Oasis & $\times$1.00 & 67.72 & 45.02 & 12.73 & 99.24 & 99.30 & 98.92 & 99.26 \\
    \rowcolor{gray!10} StableWorld + Open-Oasis & $\times$1.02 & \textbf{74.67} & \textbf{48.38} & \textbf{29.93} & \textbf{99.52} & \textbf{99.64} & 98.63 & \textbf{99.52} \\
    \midrule
    Hunyuan-GameCraft 1.0 & $\times$1.00 & 57.91 & 48.38 & 98.00 & 95.66 & 98.27 & 97.21 & 96.83 \\
    \rowcolor{gray!10} StableWorld + Hunyuan-GameCraft 1.0 & $\times$1.02 & \textbf{65.91} & \textbf{57.44} & \textbf{98.66} & \textbf{96.05} & \textbf{98.29} & 97.08 & \textbf{97.06} \\
    \bottomrule
    \end{tabular}
  }
  \label{vbench_results}
\end{table}

\begin{figure*}[t]
    \centering
    \includegraphics[width=\textwidth]{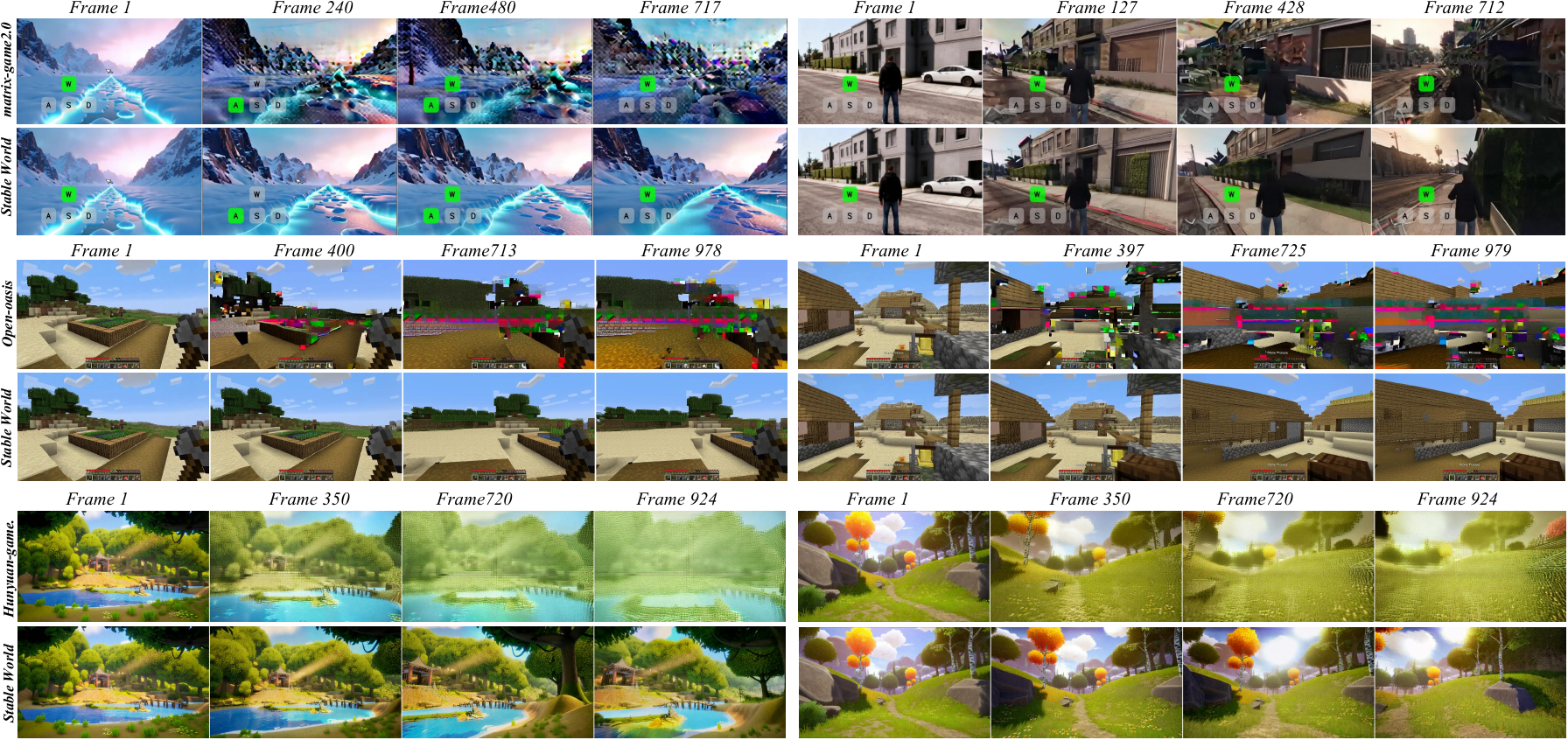}
    \caption{
    Qualitative comparison between baseline results and our proposed \textit{StableWorld} across three models.
    Each pair of rows corresponds to one model:
    (1) \textit{Matrix-Game 2.0}, 
    (2) \textit{Open-Oasis}, and 
    (3) \textit{Hunyuan-GameCraft 1.0}.
    In each group, the first row shows the baseline results and the second row shows our \textit{StableWorld} outputs, 
    which maintain higher scene stability, smoother motion, and more consistent temporal dynamics.
    }
    \label{fig:zhanshi}
\end{figure*}

\begin{figure}[htbp!]
    \centering
    \includegraphics[width=\linewidth]{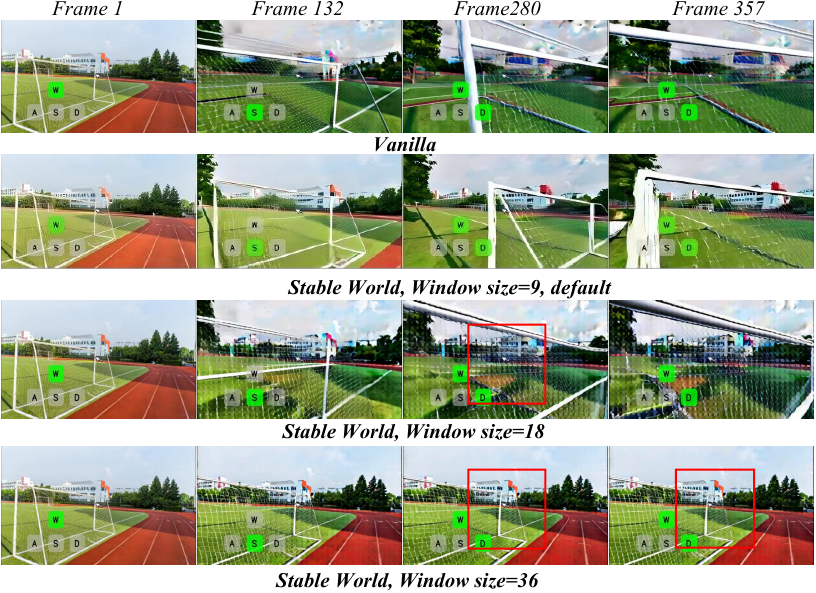}
    \caption{
    Comparison of \textit{StableWorld+Matrix-game} under different window sizes.
    }
    \label{fig:ab1}
\end{figure}


\begin{table}[t]
    \centering
    \begin{minipage}[t]{0.48\textwidth}  
        \centering
        \small
        \setlength{\tabcolsep}{4.8pt}
        \caption{User study comparing three vanilla models and StableWorld}
        \label{tab:userstudy}
        \renewcommand{\arraystretch}{1} 
        \resizebox{\linewidth}{!}{
            \begin{tabular}{lccc}
            \toprule
            \textbf{Model} & \textbf{Video Qual. ↑} & \textbf{Temporal Cons. ↑} & \textbf{Motion Smo. ↑} \\
            \midrule
            Matrix-Game 2.0 & 23.2\% & 12.5\% & 14.3\% \\
            \rowcolor{gray!10} \textbf{StableWorld + Matrix.} & \textbf{76.8\%} & \textbf{87.5\%} & \textbf{85.7\%} \\
            \midrule
            Open-Oasis & 3.6\% & 5.4\% & 7.1\% \\
            \rowcolor{gray!10} \textbf{StableWorld + Open.} & \textbf{96.4\%} & \textbf{94.6\%} & \textbf{92.9\%} \\
            \midrule
            Hunyuan-GameCraft 1.0 & 10.7\% & 3.6\% & 7.1\% \\
            \rowcolor{gray!10} \textbf{StableWorld + Hunyuan.} & \textbf{89.3\%} & \textbf{96.4\%} & \textbf{92.9\%} \\
            \bottomrule
            \end{tabular}
        }
    \end{minipage}%
    \hfill
    \begin{minipage}[t]{0.50\textwidth}  
        \centering
        \small
        \setlength{\tabcolsep}{4.8pt}
        \renewcommand{\arraystretch}{1.25} 
        \caption{Quantitative experiment: different Window size (StableWorld + MatrixGame) for performance comparison}
        \label{tab:windowstudy}
        \resizebox{\linewidth}{!}{
            \begin{tabular}{c|ccc|cc}
            \toprule
            Window size & Image Qua. ↑ & Aesthetic ↑ & Dynamic ↑ & Temporary.  ↑ & Motion. ↑ \\
            \midrule
            9(default) & 73.52  & \textbf{53.44 } & \textbf{56.66 } & \textbf{95.96 } & \textbf{98.13 } \\
            18    & 71.70  & 47.28  & 56.24  & 94.92  & 97.33  \\
            36    & \textbf{75.43 } & 52.67  & 45.52  & 95.78  & 98.36  \\
            \bottomrule
            \end{tabular}
        }
    \end{minipage}
\end{table}

\subsection{Quantitative Results}
\textbf{VBench-Long Metrics. \quad}Tab.~\ref{vbench_results} compares the VBench-Long scores between the vanilla models and our StableWorld-enhanced versions across three settings. 
We observe significant improvements in both \textit{Image Quality} and \textit{Aesthetic Quality}. 
For example, on \textit{Matrix-Game 2.0}, \textit{StableWorld} improves the Aesthetic Quality score by $14.61\%$; 
on \textit{Open-Oasis}, it improves the Image Quality score by $6.95\%$; 
and on \textit{Hunyuan-GameCraft 1.0}, it boosts the Aesthetic Quality by $9.06\%$.  
Meanwhile, we note that the \textit{Temporal Quality} and \textit{Physical Understanding} metrics do not show large differences between the vanilla models and \textit{StableWorld}. 
This is because once collapse occurs, the vanilla model tends to produce visually similar and nearly static subjects, which inadvertently stabilizes these metrics. 
Nevertheless, \textit{StableWorld} consistently achieves improvements on most metrics, 
indicating that it alleviates error accumulation while better preserving subject identity and motion continuity. Our method introduces only minimal additional computational overhead (1.00–1.02$\times$).

\noindent\textbf{User Study.\quad}Tab.~\ref{tab:userstudy} shows that \textit{StableWorld} receives the majority of votes, 
further confirming the substantial improvements in visual quality, temporal consistency, and motion smoothness achieved by \textit{StableWorld}.

\subsection{Qualitative Results}
Fig.~\ref{fig:zhanshi} visualizes qualitative results from all three models, showing that \textit{StableWorld} maintains better scene stability and reduces drift over time.
Additional qualitative comparisons are provided in Appendix~\ref{more}.
We further compare autoregressive video generation with Self-Forcing and our proposed method in Appendix~\ref{sec:more}.
These results demonstrate the generalizability and effectiveness of \textit{StableWorld}, which significantly alleviates error accumulation at its source by continuously filtering out degraded frames across different scenes.

\subsection{Ablation Studies}
\textbf{Different Window Sizes \quad} We compare different window sizes in Tab.~\ref{tab:windowstudy} and  Fig.~\ref{fig:ab1} using the \textit{Matrix-Game 2.0 + StableWorld}.  
All settings share the same action conditions.  
As the window size increases, the generation of subsequent scenes becomes affected.  
When the window size is set to 18 or 36, excessive historical frames introduce residual artifacts from previous scenes into new ones, which interfere with scene transitions. 
This issue becomes more pronounced at a window size of 36, as illustrated in Fig.~\ref{fig:ab1}. 
Consistently, the Dynamic Degree decreases significantly from 56.66 to 45.52, as shown in Tab.~\ref{tab:windowstudy}.

\begin{figure}[t]
    \centering
    \includegraphics[width=\linewidth]{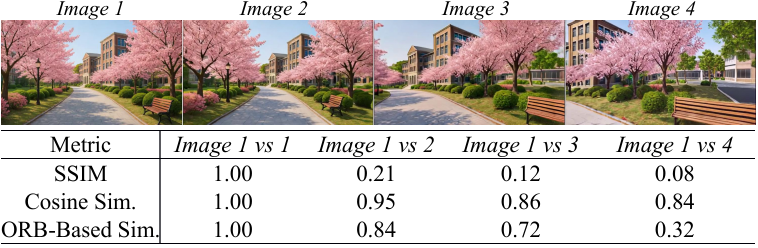}
    \caption{
    Comparison of different similarity metrics used for frame evaluation.}
    \label{fig:metric2}
\end{figure}

\begin{figure}[t]
\centering

\begin{minipage}[t]{0.49\textwidth}
    \centering
    \includegraphics[width=\linewidth]{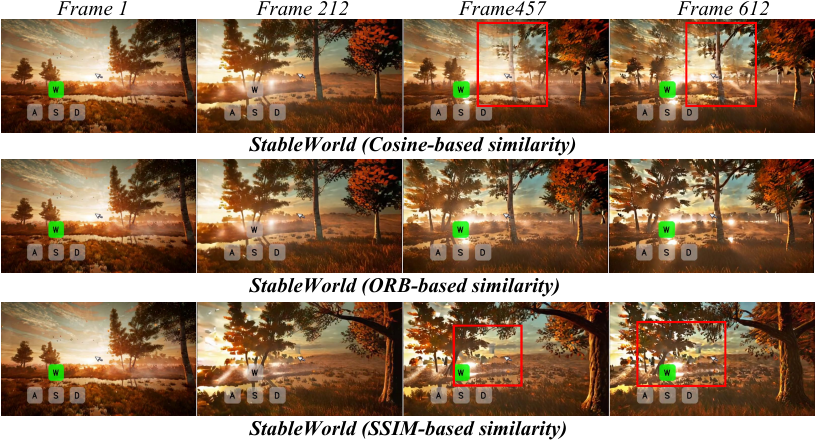}
    
    \caption{ Qualitative comparison of \textit{StableWorld + MatrixGame} under different similarity metrics :
    Cosine-based similarity,
    ORB-based similarity
    and SSIM-based similarity.}
    \label{fig:sim_matrix_metrics}
\end{minipage}
\hfill
\begin{minipage}[t]{0.48\textwidth}
    \centering
    \includegraphics[width=\linewidth]{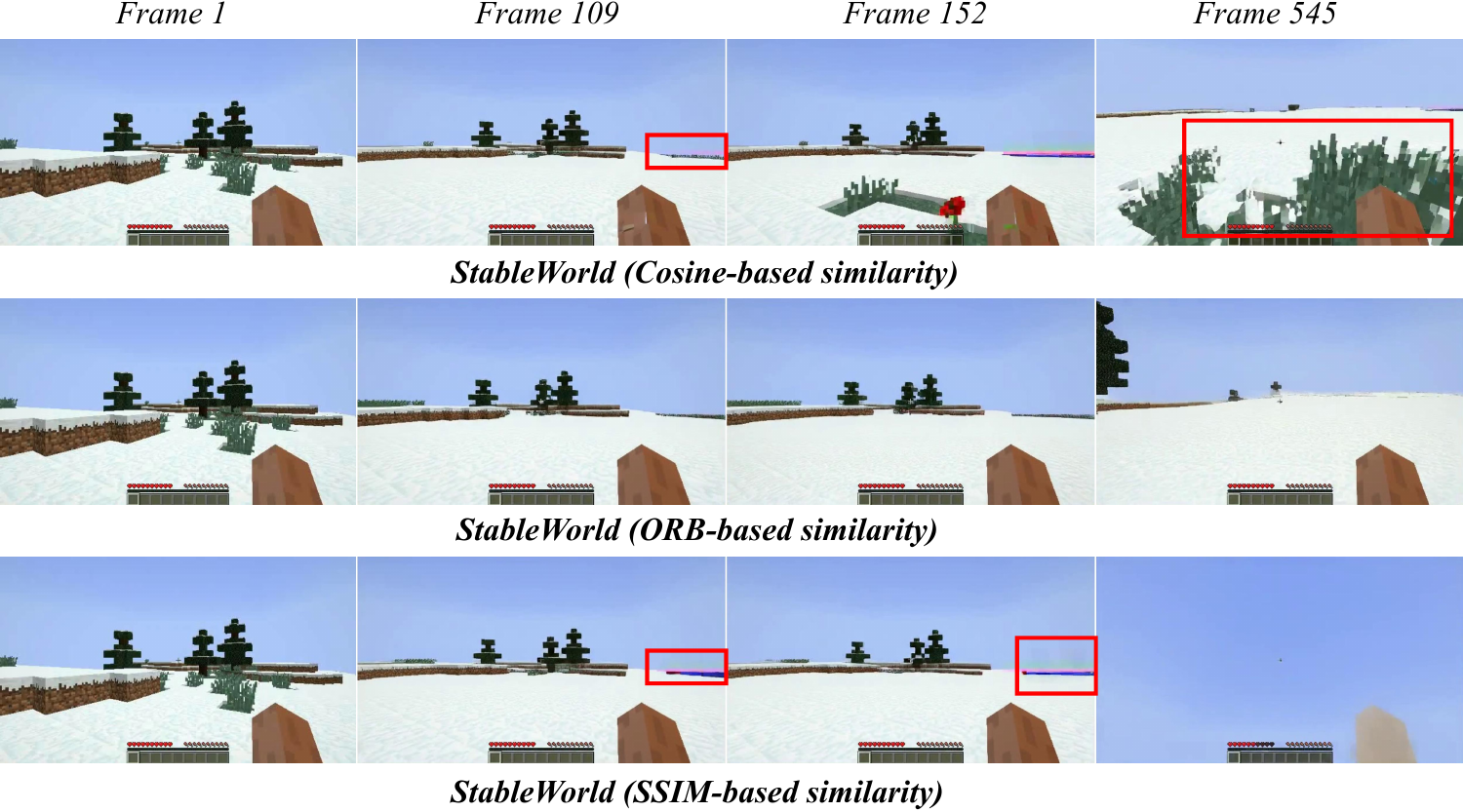}
    \caption{ Qualitative comparison of \textit{StableWorld + Open-Oasis} under different similarity metrics:
    Cosine-based similarity,
    ORB-based similarity
    and SSIM-based similarity.}
    \label{fig:sim_oasis_metrics}
\end{minipage}

\end{figure}

\begin{table}[htbp!]
\centering
\small
\setlength{\tabcolsep}{4.5pt}
\renewcommand{\arraystretch}{1.15}
\caption{
Ablation study on different similarity metrics under 
\textit{StableWorld + MatrixGame}. 
ORB-based similarity achieves the strongest overall performance across visual and temporal metrics.
}
\vspace{2pt}
\resizebox{\linewidth}{!}{
\begin{tabular}{c|ccc|cc|cc}
\toprule
\textbf{Similarity} 
& \textbf{Image Qual. ↑} 
& \textbf{Aesthetic ↑} 
& \textbf{Dynamic Deg. ↑} 
& \textbf{Temp. Flick. ↑} 
& \textbf{Motion Smooth ↑} 
& \textbf{Subj. Cons. ↑} 
& \textbf{Bg. Cons. ↑} \\
\midrule
SSIM 
& 69.52 
& 50.53 
& 56.66
& 94.64 
& 97.33 
& 96.53 
& 97.14 \\

Cosine Sim. 
& 70.16 
& 48.78 
& 55.29 
& 94.80 
& 97.98 
& \textbf{97.28} 
& 97.72 \\

\rowcolor{gray!12}
ORB-Based Sim. 
& \textbf{73.52} 
& \textbf{53.44} 
& \textbf{56.66} 
& \textbf{95.96} 
& \textbf{98.13} 
& 97.02 
& \textbf{97.74} \\
\bottomrule
\end{tabular}
}
\label{tab:similarity_metrics}
\vspace{-4pt}
\end{table}

\begin{table}[htbp!]
\centering
\small
\setlength{\tabcolsep}{4.5pt}
\renewcommand{\arraystretch}{1.15}
\caption{
Ablation study on different similarity metrics under 
\textit{StableWorld + Open-oasis}. 
ORB-based similarity achieves the strongest overall performance across visual and temporal metrics.
}
\vspace{2pt}
\resizebox{\linewidth}{!}{
\begin{tabular}{l|ccccccc}
\toprule
\textbf{Similarity} 
& \textbf{Image Qual. ↑} 
& \textbf{Aesthetic ↑} 
& \textbf{Dynamic Deg. ↑} 
& \textbf{Temp. Flick. ↑} 
& \textbf{Motion Smooth ↑} 
& \textbf{Subj. Cons. ↑} 
& \textbf{Bg. Cons. ↑} \\
\midrule
SSIM 
& 72.29 
& 47.51 
& \textbf{29.93} 
& \textbf{99.69} 
& 99.49 
& 97.79 
& 99.51 \\

Cosine Sim. 
& 72.94 
& 47.47 
& 26.38 
& 99.38 
& 99.20 
& 97.56 
& 99.12 \\

\rowcolor{gray!12}
ORB-Based Sim. 
& \textbf{74.67} 
& \textbf{48.38} 
& \textbf{29.93} 
& 99.52 
& \textbf{99.64} 
& \textbf{98.63} 
& \textbf{99.52} \\
\bottomrule
\end{tabular}
}
\label{tab:similarity_metrics_oasis}
\vspace{-4pt}
\end{table}

\noindent\textbf{Different Similarity Metrics. \quad} 
We further compare different similarity metrics, including SSIM and cosine similarity. 
We conduct experiments on both \textit{Matrix-Game 2.0 + StableWorld} and \textit{Open-Oasis + StableWorld}. 
The qualitative results are shown in Fig.~\ref{fig:sim_matrix_metrics} and Fig.~\ref{fig:sim_oasis_metrics}.  
The SSIM threshold is set to 0.3, and the cosine similarity threshold is set to 0.8.  
SSIM is overly sensitive to geometric perspective changes, while cosine similarity is less sensitive to spatial transformations, as also illustrated in Fig.~\ref{fig:metric2}.  
As a result, cosine-based similarity sometimes fails to detect scene changes, causing frames from old scenes to remain in the window, which leads to visible artifacts in the newly generated scene.  
In contrast, SSIM tends to evict clean frames too early, resulting in more severe cumulative errors compared to other metrics.  
The quantitative results are reported in Tab.~\ref{tab:similarity_metrics} and Tab.~\ref{tab:similarity_metrics_oasis}, which further demonstrate the superiority of ORB-based similarity over other similarity metrics.

\begin{figure}[htbp!]
\centering
\begin{minipage}[t]{0.45\textwidth}
    \centering
    \includegraphics[width=\linewidth]{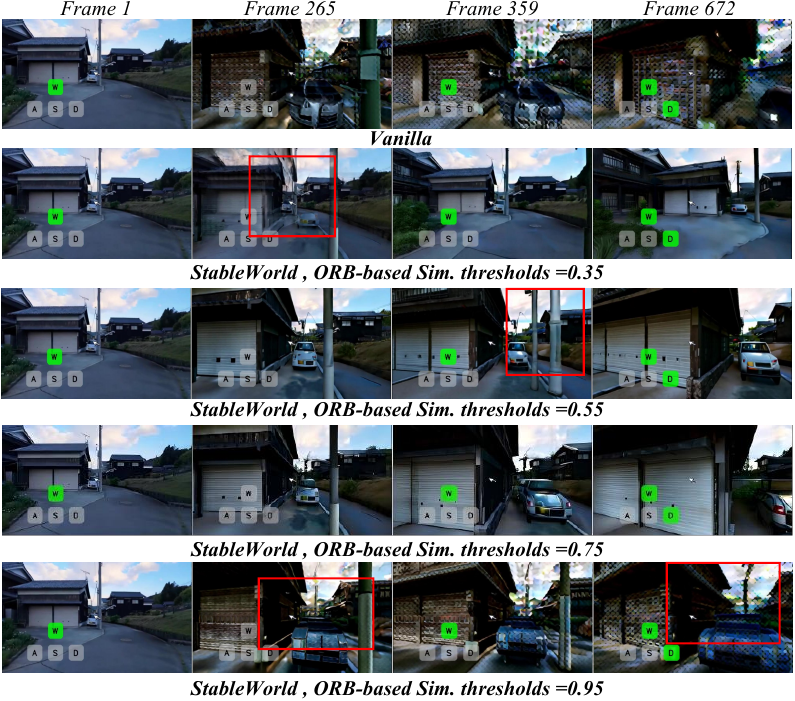}
    \caption{
    Qualitative comparison of \textit{StableWorld + MatrixGame} under four ORB-based similarity thresholds (0.35, 0.55, 0.75, 0.95).
    }
    \label{fig:sim_matrix}
\end{minipage}
\hfill
\begin{minipage}[t]{0.515\textwidth}
    \centering
    \includegraphics[width=\linewidth]{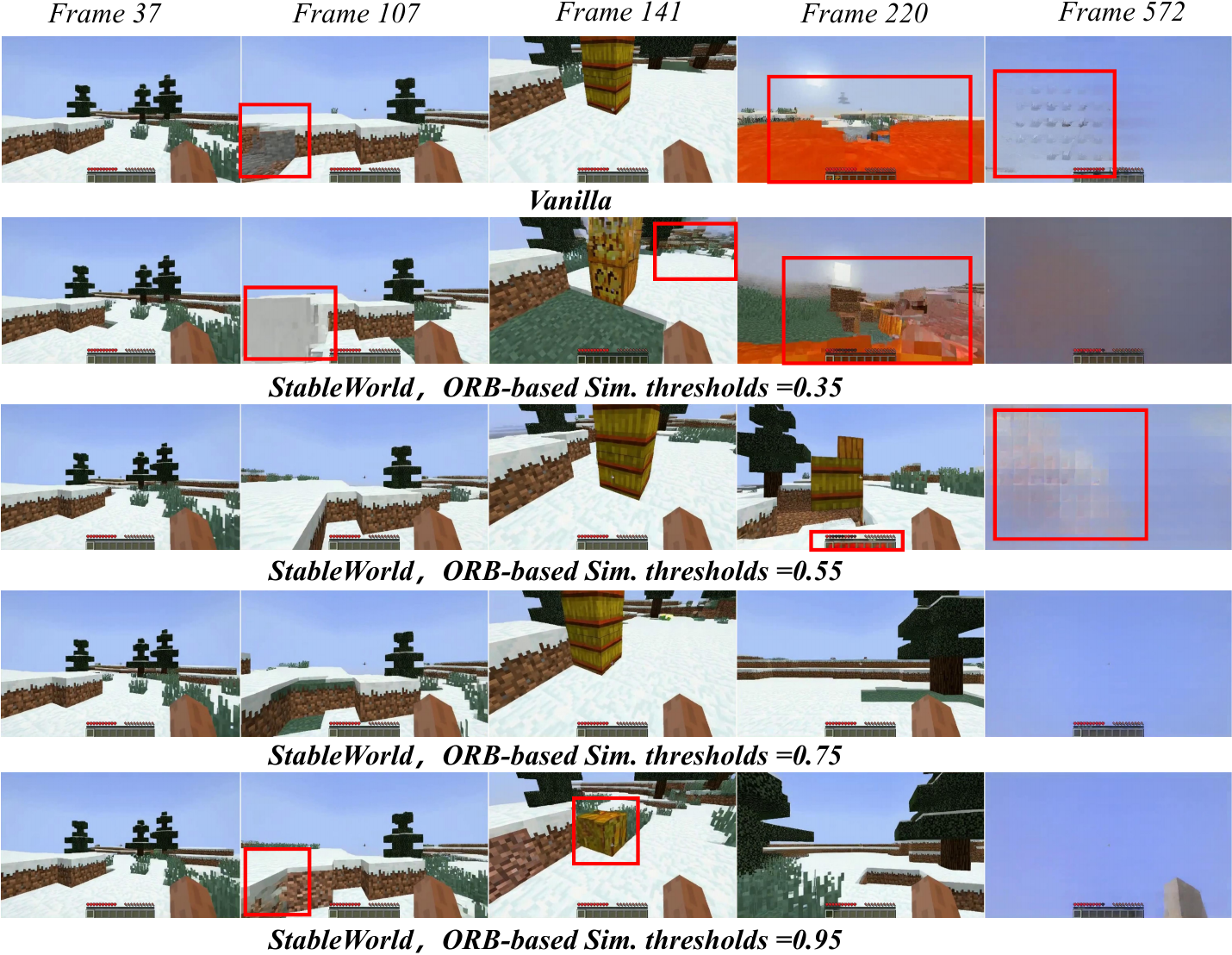}
    \caption{
        Qualitative comparison of \textit{StableWorld + Open-Oasis} under four ORB-based similarity thresholds (0.35, 0.55, 0.75, 0.95).
    }
    \label{fig:sim_oasis}
\end{minipage}
\end{figure}

\begin{table}[htbp!]
\centering
\small
\setlength{\tabcolsep}{4.5pt}
\renewcommand{\arraystretch}{1.15}
\caption{
Ablation study on ORB-based similarity thresholds under 
\textit{StableWorld + MatrixGame}. 
The default threshold (0.75) achieves the best trade-off between
visual quality and temporal stability.
}
\vspace{2pt}
\resizebox{\linewidth}{!}{
\begin{tabular}{c|ccc|cc|cc}
\toprule
\textbf{ORB-Thr.} 
& \textbf{Image Qual. ↑} 
& \textbf{Aesthetic ↑} 
& \textbf{Dynamic Deg. ↑} 
& \textbf{Temp. Flick. ↑} 
& \textbf{Motion Smooth ↑} 
& \textbf{Subj. Cons. ↑} 
& \textbf{Bg. Cons. ↑} \\
\midrule
0.35 & 71.50 & 49.64 & 56.19 & 94.66 & 97.19 & 97.21 & 97.02 \\
0.55 & 71.65 & 49.73 & 56.26 & 94.66 & 97.20 & \textbf{97.26} & 97.13 \\
\rowcolor{gray!12}
\textbf{0.75 (default)} 
& \textbf{73.52} 
& \textbf{53.44} 
& \textbf{56.66} 
& \textbf{95.96} 
& \textbf{98.13} 
& 97.02 
& \textbf{97.74} \\
0.95 & 69.23 & 51.65 & \textbf{56.66} & 94.75 & 97.72 & 96.37 & 96.73 \\
\bottomrule
\end{tabular}
}
\label{tab:orb_threshold}
\end{table}

\begin{table}[htbp!]
\centering
\small
\setlength{\tabcolsep}{4.5pt}
\renewcommand{\arraystretch}{1.15}
\caption{
Ablation study on ORB-based similarity thresholds under 
\textit{StableWorld + Open-Oasis}. 
The default threshold (0.75) achieves the best trade-off between visual quality and temporal stability.
}
\vspace{2pt}
\resizebox{\linewidth}{!}{
\begin{tabular}{c|ccc|cc|cc}
\toprule
\textbf{ORB-Thr.} 
& \textbf{Image Qual. ↑} 
& \textbf{Aesthetic ↑} 
& \textbf{Dynamic Deg. ↑} 
& \textbf{Temp. Flick. ↑} 
& \textbf{Motion Smooth ↑} 
& \textbf{Subj. Cons. ↑} 
& \textbf{Bg. Cons. ↑} \\
\midrule

0.35 
& 73.28 
& 48.24 
& 28.96 
& 97.48 
& 97.86 
& 91.27 
& 97.60 \\

0.55 
& 66.26 
& 40.32 
& 27.19
& 97.74 
& 98.05 
& 92.49 
& 96.60 \\

\rowcolor{gray!12}
\textbf{0.75 (default)} 
& \textbf{74.67} 
& 48.38 
& \textbf{29.93} 
& \textbf{99.52} 
& \textbf{99.64} 
& \textbf{98.63} 
& 99.52 \\

0.95 
& 74.52 
& \textbf{49.60} 
& 13.16 
& 99.10 
& 99.25 
& 97.89 
& \textbf{99.55} \\

\bottomrule
\end{tabular}
}
\label{tab:orb_threshold1}
\vspace{-4pt}
\end{table}

\noindent\textbf{Different ORB-Based Similarity Threshold. \quad}
We further evaluate the impact of different ORB-based similarity thresholds (0.35, 0.55, 0.75, 0.95) 
using \textit{Matrix-Game 2.0 + StableWorld} and \textit{Open-Oasis + StableWorld}. 
The qualitative results are shown in Fig.~\ref{fig:sim_matrix} and Fig.~\ref{fig:sim_oasis}.  
A lower threshold (e.g., 0.35) retains old frames for too long, hindering the generation of new scenes.  
Similarly, a threshold of 0.55 still causes noticeable interference between adjacent scenes.  
In general, lower thresholds tend to introduce significant visual artifacts.  
In contrast, a very high threshold (e.g., 0.95) leads to the premature eviction of clean frames, introducing mild cumulative errors.  
Overall, the threshold of 0.75 achieves the best trade-off, effectively reducing error accumulation without compromising motion consistency.  
The quantitative results shown in Tab.~\ref{tab:orb_threshold} and Tab.~\ref{tab:orb_threshold1} further confirm that the default threshold (0.75) achieves the best balance between visual quality and temporal stability.

%% file: sec/6_appendix.tex
\clearpage
\setcounter{page}{1}

\section*{Appendix} 
\appendix




\section{Algorithmic Implementation}
\label{Alg}

\begin{algorithm}[h]
\caption{Dynamic Frame Eviction via ORB-based Geometric Similarity}
\label{alg:orb_eviction}
\KwIn{Sliding window $\{\boldsymbol{L}_0, \ldots, \boldsymbol{L}_{N-1}\}$ in latent space; pixel frames $\{{P}_0, \ldots, {P}_{N-1}\}$,  earlier frames in the window $\{P_0, P_1, \ldots, P_K\}$ with $K < N-1$; similarity threshold $\theta$.}
\KwOut{Updated window after eviction.}

Extract ORB features from the reference frame ${P}_0$. \\[2pt]
\For{$k = 1$ \KwTo $K$}{
    Extract ORB features from the current frame ${P}_k$. \\
    Compute descriptor matches 
    $G = \{(i,j) \mid \| \boldsymbol{d}_i^{(0)} - \boldsymbol{d}_j^{(k)} \| < \tau \}$. \\
    Estimate Homography $\mathbf{H}$ and Fundamental matrix $\mathbf{F}$ via RANSAC. \\
    Compute inlier ratios 
    $r_H = \frac{|\mathcal{I}_{H}|}{|G|}$, 
    $r_F = \frac{|\mathcal{I}_{F}|}{|G|}$, 
    and define similarity 
    $s({P}_0, {P}_k) = \max(r_H, r_F)$. \\[2pt]
    \If{$s({P}_0, {P}_k) < \theta$}{
        \textbf{break}
    }
}
Evict the farthest latent frame ${P}_{k-1}$ from the window. \\
Insert the newly generated frame to maintain window size $N$.
\end{algorithm}

In practice, we implement the geometric similarity computation in Algorithm~\ref{alg:orb_eviction} using an ORB-based matching procedure. 
For each frame pair $({P}_0, {P}_k)$, up to 3000 ORB keypoints are extracted with OpenCV’s \texttt{ORB\_create} (FAST threshold = 7). 
Descriptor matching is performed using a brute-force matcher with Hamming distance and a two-nearest-neighbor ratio test. 
The ratio threshold $\tau$ is set to 0.8 following Lowe’s criterion, and the resulting good matches are further filtered by RANSAC using both the Homography $\mathbf{H}$ and Fundamental matrix $\mathbf{F}$ models. 
The RANSAC reprojection tolerance $\epsilon$ is fixed to 3.0 pixels, and the confidence level is set to 0.99 for Fundamental-matrix estimation.
The inlier ratios $r_H$ and $r_F$ are computed as the proportions of the inlier correspondence sets $\mathcal{I}_H$ and $\mathcal{I}_F$ over all valid matches, and the final similarity score $s({P}_0, {P}_k)$ is defined as $\max(r_H, r_F)$. 
Frames with fewer than five valid correspondences are considered unreliable and assigned a similarity score of $0$. 
We empirically set the eviction threshold $\theta$ to $0.75$ in all experiments.

\section{Additional Ablation}
\label{abl}

\begin{figure}[t]
    \centering
    \includegraphics[width=\linewidth]{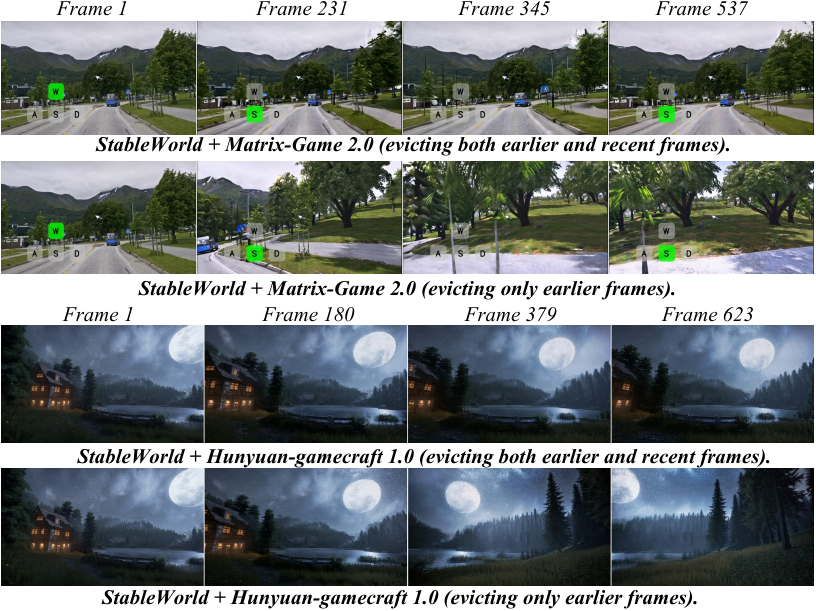}
    \caption{
    Comparison between evicting both earlier and recent frames  and evicting only earlier frames in \textit{StableWorld}.
    }
    \label{fig:df}
\end{figure}

\textbf{Length of Frames Evicted in the Window.} 
We further investigate the effect of evicting both earlier frames and recent frames, as illustrated in Fig.~\ref{fig:df}. 
All settings are tested under identical action conditions for fair comparison. 
As shown in the figure, evicting recent frames leads to inconsistent motion and unstable short-term dynamics, which in turn hinders the generation of subsequent frames. 
This behavior is analogous to retaining too many frames in the window: both strategies reduce the model's adaptability to scene changes and limit its ability to transition smoothly to new environments. 
In contrast, selectively removing earlier degraded frames preserves local motion continuity while maintaining sufficient flexibility for handling scene transitions.

\section{More Qualitative Results}
\label{more}

\begin{figure*}[t]
    \centering
    \includegraphics[width=\linewidth]{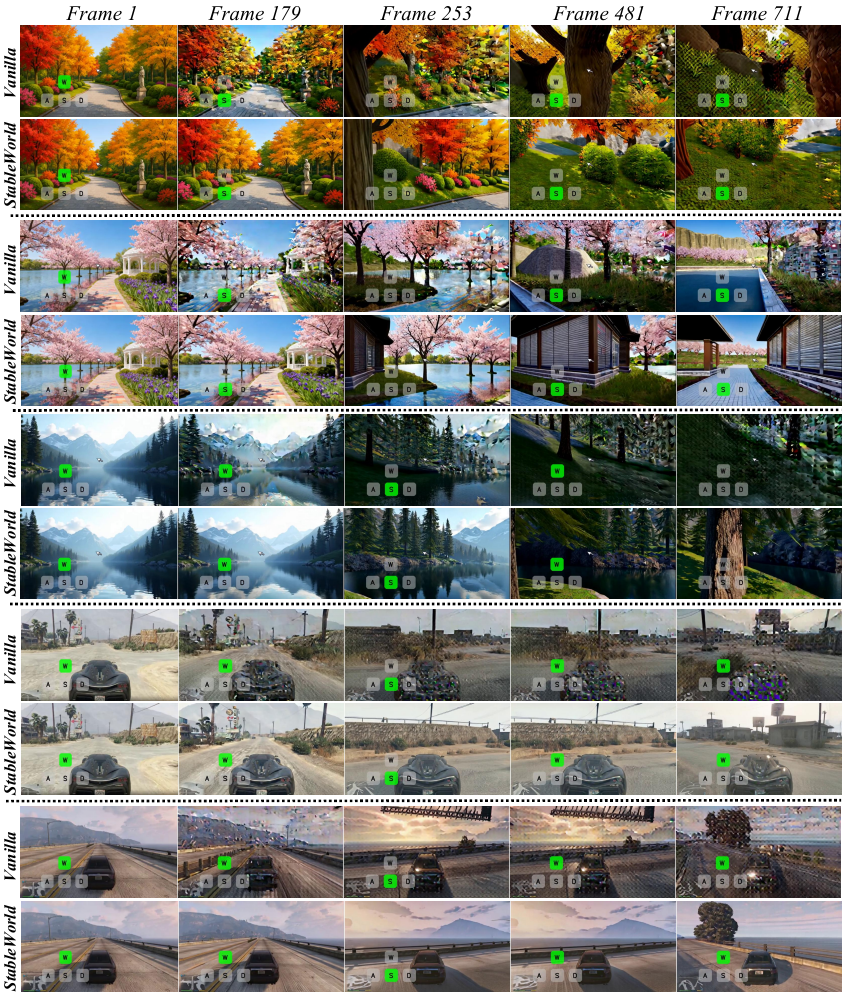}
    \caption{
    Additional qualitative comparison between \textit{Matrix-Game 2.0} and our \textit{StableWorld}. 
    The first row shows \textit{Matrix-Game 2.0}; the second row shows \textit{StableWorld}. 
    \textit{StableWorld} shows more stable results with better visual quality.
    }
    \label{fig:matrix1}
\end{figure*}

\begin{figure*}[t]
    \centering
    \includegraphics[width=\linewidth]{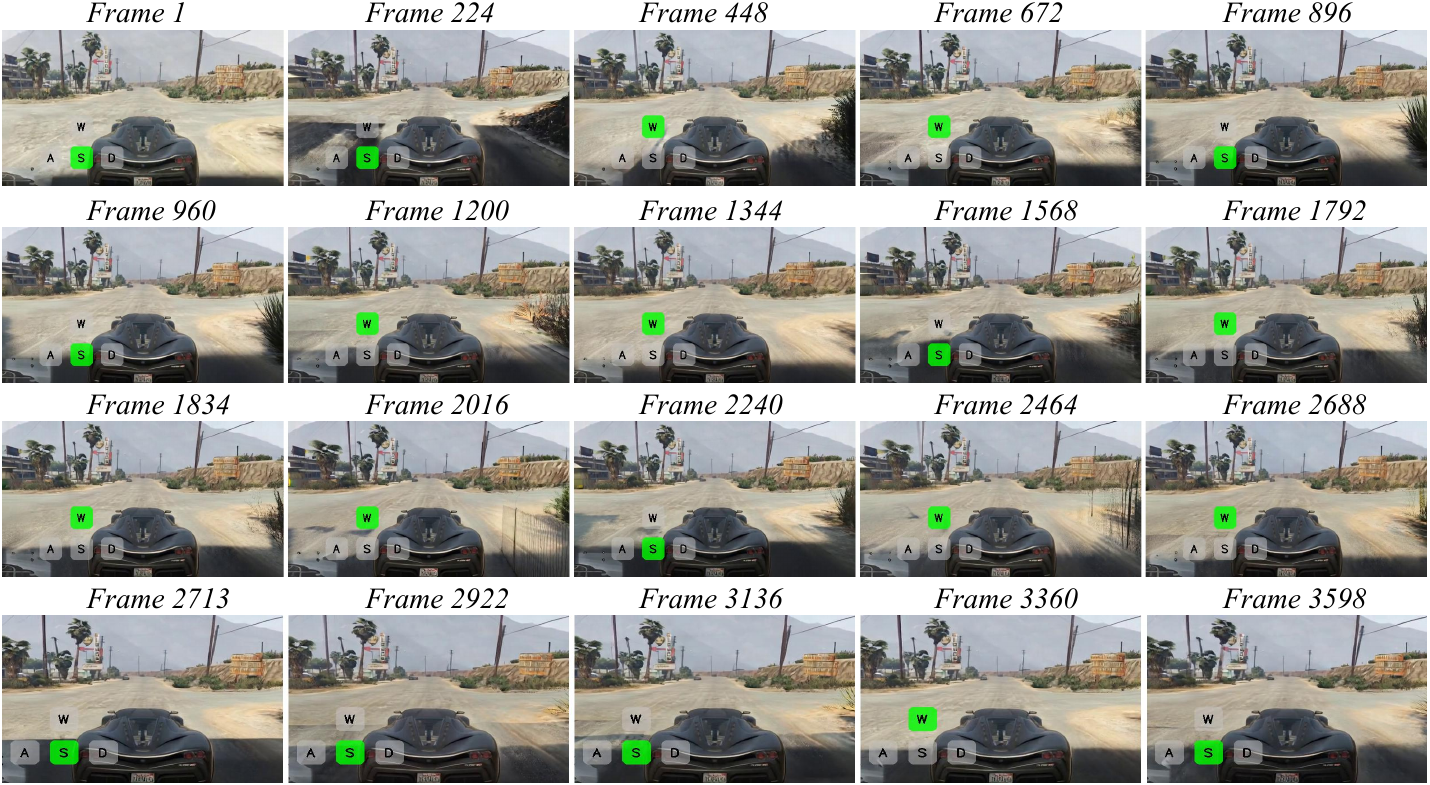}
\caption{
Long-horizon generation results of \textit{StableWorld} under small-motion scenarios. 
The model maintains scene stability over thousands of frames without  drift or degradation.
}
    \label{fig:stay}
\end{figure*}

\begin{figure*}[t]
    \centering
    \includegraphics[width=\linewidth]{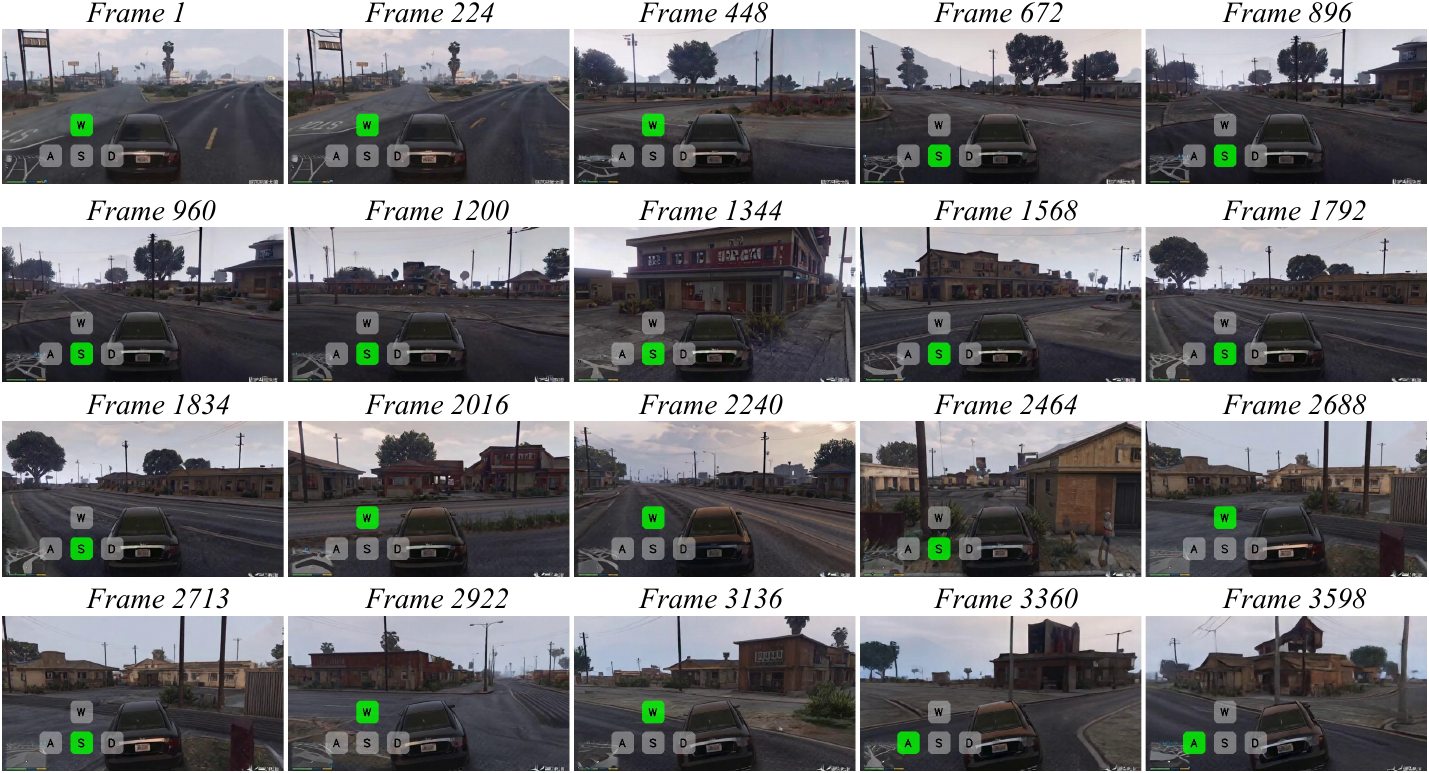}
\caption{
Long-horizon generation results of \textit{StableWorld} under large-motion scenarios. 
Despite significant viewpoint and motion changes, \textit{StableWorld} preserves temporal consistency and avoids cumulative drift.
}
    \label{fig:motion}
\end{figure*}

\begin{figure*}[t]
    \centering
    \includegraphics[width=\linewidth]{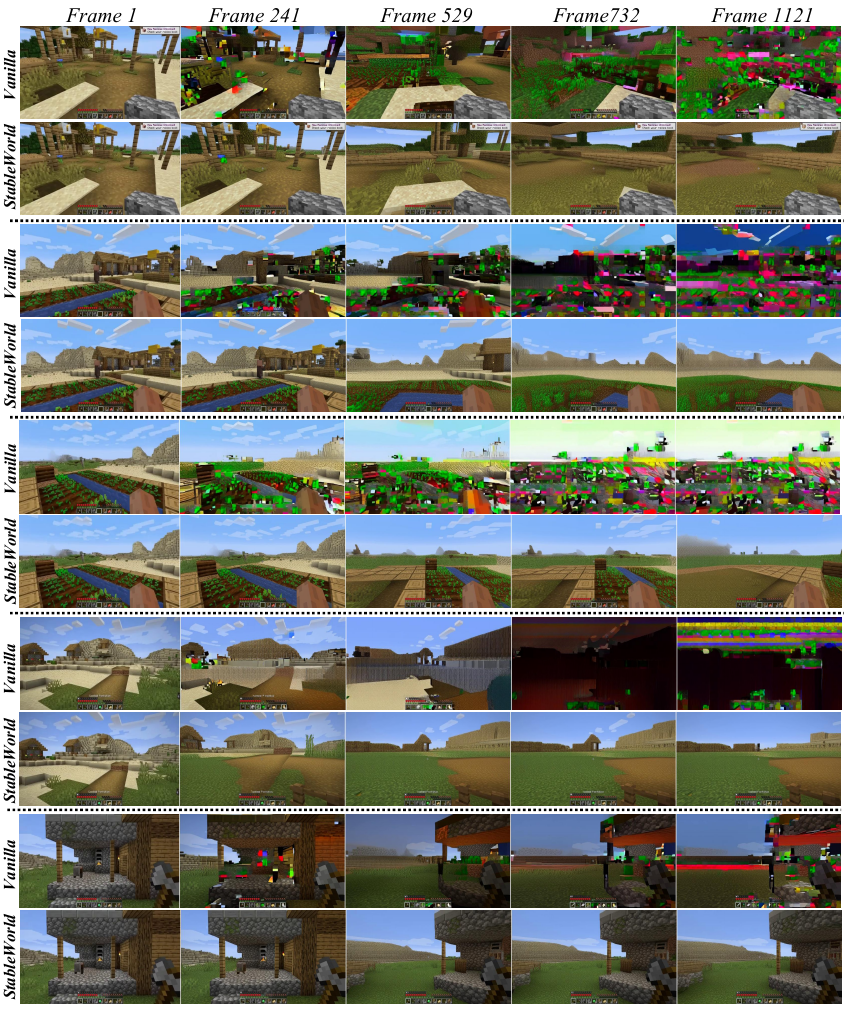}
    \caption{
    Additional qualitative comparison between \textit{Open-Oasis} and our \textit{StableWorld}. 
    The first row shows \textit{Open-Oasis}; the second row shows \textit{StableWorld}. 
    \textit{StableWorld} shows more stable results with better visual quality.
    }
    \label{fig:oasis1}
\end{figure*}

\begin{figure*}[t]
    \centering
    \includegraphics[width=\linewidth]{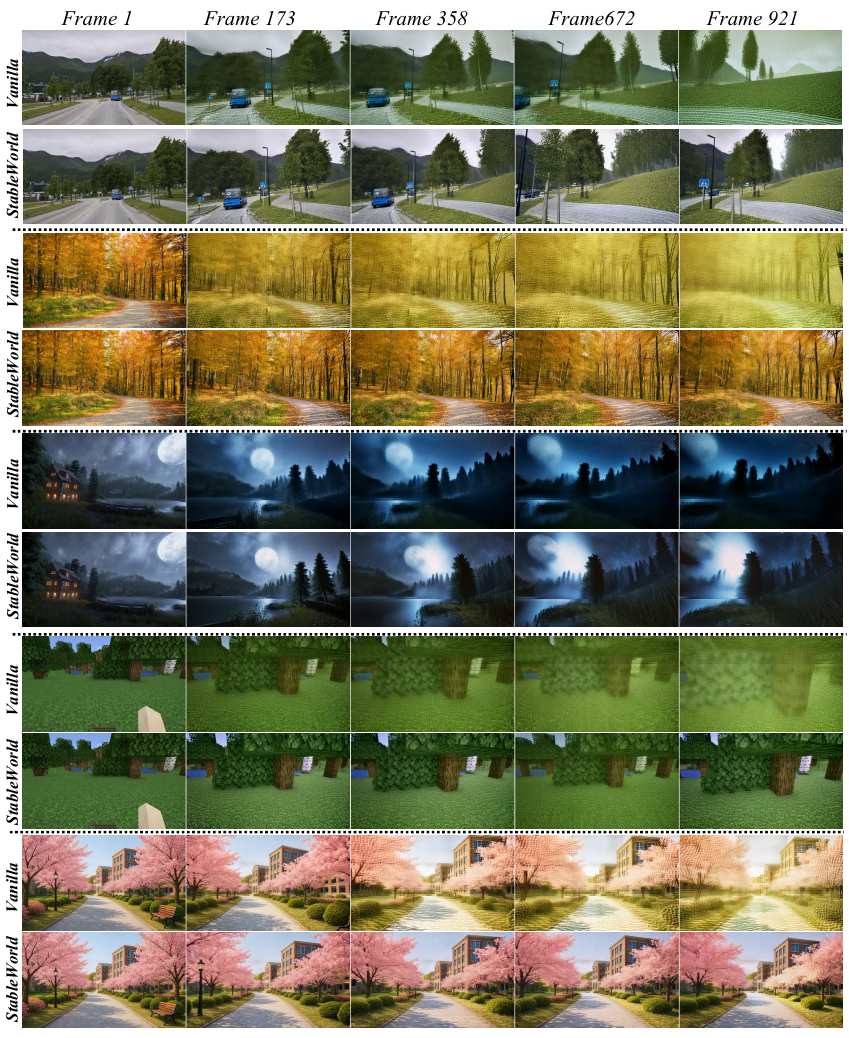}
    \caption{
    Additional qualitative comparison between \textit{Hunyuan-GameCraft 1.0} and our \textit{StableWorld}. 
    The first row shows \textit{Hunyuan-GameCraft 1.0}; the second row shows \textit{StableWorld}. 
    \textit{StableWorld} shows more stable results with better visual quality.
    }
    \label{fig:hunyuan1}
\end{figure*}

We present additional qualitative results for \textit{Matrix-Game 2.0} in Fig.~\ref{fig:matrix1}. From these more diverse outputs, we observe that \textit{StableWorld} significantly strengthens stability during long-horizon generation. 
We further evaluate \textit{StableWorld} on extremely long frame sequences (e.g., thousands of frames) under both small-motion settings (Fig.~\ref{fig:stay}) and large-motion scenarios (Fig.~\ref{fig:motion}). More Qualitative comparisons on \textit{Open-Oasis} and \textit{Hunyuan-GameCraft 1.0} are provided in Fig.~\ref{fig:oasis1} and Fig.~\ref{fig:hunyuan1}, respectively.
Across all settings, we observe that our method effectively prevents cumulative errors by continuously filtering out degraded frames while maintaining coherent motion. As a result, \textit{StableWorld} produces notably more stable and temporally consistent interactive video sequences.

\section{Qualitative Results in Autoregressive Video Generation}
\label{sec:more}

We further validate the effectiveness of \textit{StableWorld} in long-horizon autoregressive video generation. We use the Self-Forcing model as our baseline.

\begin{figure}[t]
    \centering
    \includegraphics[width=\linewidth]{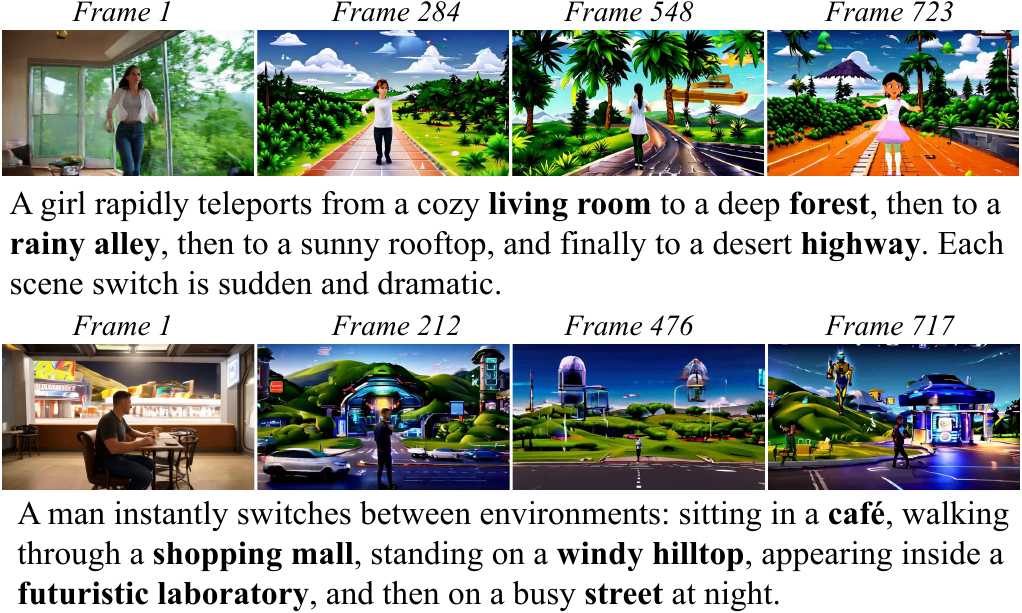}
\caption{
Qualitative results of \textit{Self-Forcing} in a fast-changing scene. 
When the scene changes frequently, the model shows fewer cases of scene collapse, 
although the visuals may still appear inconsistent across frames.
}
    \label{fig:selfmotion}
\end{figure}

\begin{figure}[t]
    \centering
    \includegraphics[width=\linewidth]{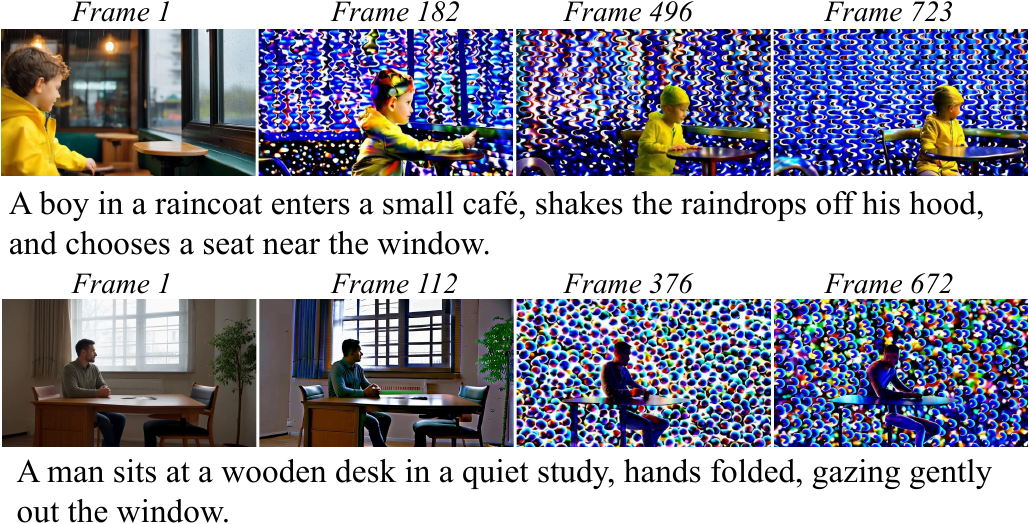}
\caption{
Qualitative results of \textit{Self-Forcing} in a low-motion scene. 
Even with minimal scene changes, Self-Forcing exhibits noticeable drift accumulation that eventually leads to scene collapse.
}
    \label{fig:selfstay}
\end{figure}

We also observe that Self-Forcing exhibits the same phenomenon as in the interactive video generation setting. 
As shown in Fig.~\ref{fig:selfmotion}, when the scene changes frequently, catastrophic collapse rarely occurs.
In contrast, in Fig.~\ref{fig:selfstay}, where the model stays within the same scene and the changes are relatively small, collapse emerges clearly: small inter-frame drift gradually accumulates within the same scene and eventually leads to severe degradation. 
This confirms that error accumulation in autoregressive generation is closely linked to drift propagation within a persistent scene.

\begin{figure*}[t]
    \centering
    \includegraphics[width=\linewidth]{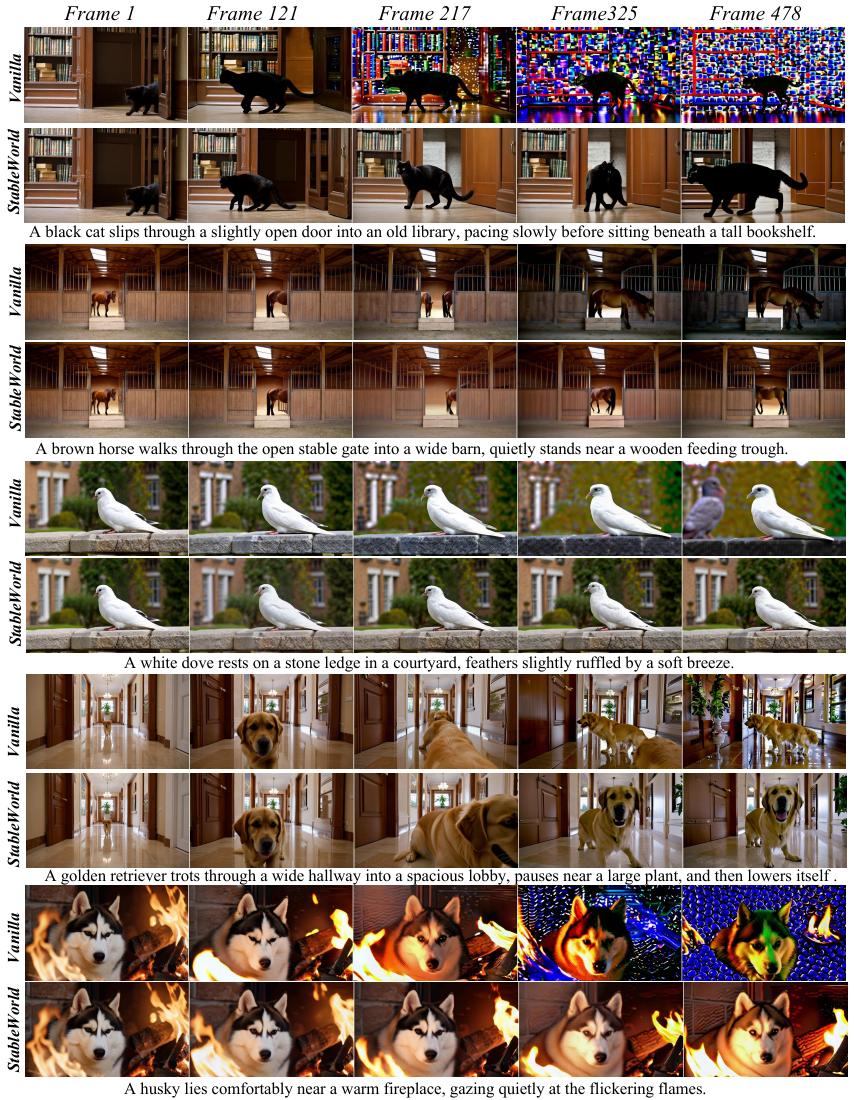}
    \caption{
    Additional qualitative comparison between \textit{self-forcing} and our \textit{StableWorld}.  
    \textit{StableWorld} shows more stable results with better visual quality.
    }
    \label{fig:self}
\end{figure*}

We then integrate our method into the Self-Forcing framework. 
We set the key–value cache length to $9$ and treat the earliest $6$ frames as reference frames. 
Since the model produces $3$ frames per step, we compute the ORB-based geometric similarity between the 3rd and 6th frames to decide whether the earliest frames should be evicted. 
A qualitative comparison is shown in Fig.~\ref{fig:self}.  
As can be seen, \textit{StableWorld} significantly alleviates error accumulation in Self-Forcing, resulting in more stable, consistent, and higher-quality long videos.  
These results demonstrate that our approach offers a promising solution for mitigating drift in long-term autoregressive video generation.

\section{Limitation}
\label{sec:lim}

One limitation of our method is the slight increase in inference time introduced by the computation of ORB-based geometric similarity, resulting in approximately a 1.01$\times$--1.02$\times$ slowdown compared with the vanilla model under the default setting. However, we believe this enables the sliding-window length to become dynamically adjustable by discarding drifted frames, thereby reducing its associated computational cost, which in turn lowers the overall inference time. We leave this direction for future exploration.


